\documentclass[10pt]{article}

\PassOptionsToPackage{natbib=true}{biblatex}
\PassOptionsToPackage{verbatim}{minted}
\usepackage{aithyra}
\usepackage{flafter}
\usepackage{float}
\usepackage{placeins}
\usepackage{longtable}
\usepackage{pgfplots}
\pgfplotsset{compat=1.18}
\newcommand{\na}{---}
\newcommand{\best}[1]{\textbf{#1}}
\newcommand{\second}[1]{\underline{#1}}
\newcommand{\third}[1]{#1}
\colorlet{secrow}{gray!13}

\definecolor{galleryfill}{HTML}{F7F7F4}
\definecolor{galleryborder}{HTML}{D4D4CF}
\definecolor{gallerymuted}{HTML}{66665F}
\newsavebox{\galleryleft}
\newsavebox{\galleryright}
\newlength{\gallerywidth}
\newlength{\galleryheight}
\newcommand{\galleryfontsize}{8.2}
\newcommand{\galleryleading}{9.4}
\DeclareMathSizes{8.2}{8.2}{6}{5}
\DeclareMathSizes{7.3}{7.3}{5}{5}
\newcommand{\gallerytoken}[1]{{\color{gallerymuted}\ttfamily\fontsize{7.3}{8.6}\selectfont #1}}
\newcommand{\galleryurl}[1]{{\urlstyle{same}\nolinkurl{#1}}}
\newcommand{\gallerycard}[4]{%
  \begin{minipage}[t]{\gallerywidth}
  \fontsize{\galleryfontsize}{\galleryleading}\selectfont\raggedright
  \setlength{\parindent}{0pt}\setlength{\parskip}{0pt}%
  \setlength{\emergencystretch}{1em}%
  \noindent\textbf{#1}\hfill
  {\color{gallerymuted}\sffamily\fontsize{7.3}{8.6}\selectfont\##3}\par
  {\color{gallerymuted}\sffamily\fontsize{7.3}{8.6}\selectfont\strut #2\par}
  \vspace{4pt}%
  #4\par
  \end{minipage}%
}
\newcommand{\galleryframe}[1]{%
  \fcolorbox{galleryborder}{galleryfill}{%
    \usebox{#1}}%
}
\newcommand{\gallerypair}[2]{%
  \begingroup
  \setlength{\parskip}{0pt}%
  \par\addvspace{6pt}%
  \setlength{\fboxsep}{4pt}\setlength{\fboxrule}{0.35pt}%
  \setlength{\gallerywidth}{\dimexpr(\linewidth-12pt)/2-2\fboxsep-2\fboxrule\relax}%
  \sbox{\galleryleft}{\ignorespaces#1\unskip}%
  \sbox{\galleryright}{\ignorespaces#2\unskip}%
  \setlength{\galleryheight}{\dimexpr\ht\galleryleft+\dp\galleryleft\relax}%
  \ifdim\dimexpr\ht\galleryright+\dp\galleryright\relax>\galleryheight
    \setlength{\galleryheight}{\dimexpr\ht\galleryright+\dp\galleryright\relax}%
  \fi
  \noindent\hbox to\linewidth{%
    \galleryframe{\galleryleft}\hfil
    \ifdim\wd\galleryright>0pt\galleryframe{\galleryright}\fi}%
  \par\endgroup
}

\newcommand{\gallerysingle}[1]{%
  \par\addvspace{6pt}%
  \begingroup
  \setlength{\fboxsep}{4pt}\setlength{\fboxrule}{0.35pt}%
  \setlength{\gallerywidth}{\dimexpr\linewidth-2\fboxsep-2\fboxrule\relax}%
  \sbox{\galleryleft}{\ignorespaces#1\unskip}%
  \noindent\galleryframe{\galleryleft}\par
  \endgroup
}

\hypersetup{
  pdftitle={Hacking Generative Perplexity: Why Unconditional Text Evaluation Needs New Metrics},
  pdfauthor={Antonio Franca and Alexander Tong},
  pdfsubject={AITHYRA preprint},
  pdfkeywords={generative perplexity, evaluation, diffusion language models, distributional metrics, compression}
}

\newcommand{\genppl}{\text{gen-PPL}}
\newcommand{\Hemp}{H_{\mathrm{uni}}}
\newcommand{\Hpos}{H_{\mathrm{pos}}}

\title{Hacking Generative Perplexity: Why Unconditional Text Evaluation Needs New Metrics}

\author{Antonio Franca\qquad Alexander Tong \\
AITHYRA, Vienna, Austria \\
\texttt{afranca@aithyra.ac.at}}

\date{22 October, 2026}

\makeatletter
\renewcommand{\maketitle}{%
  \noindent\includegraphics[height=0.4in]{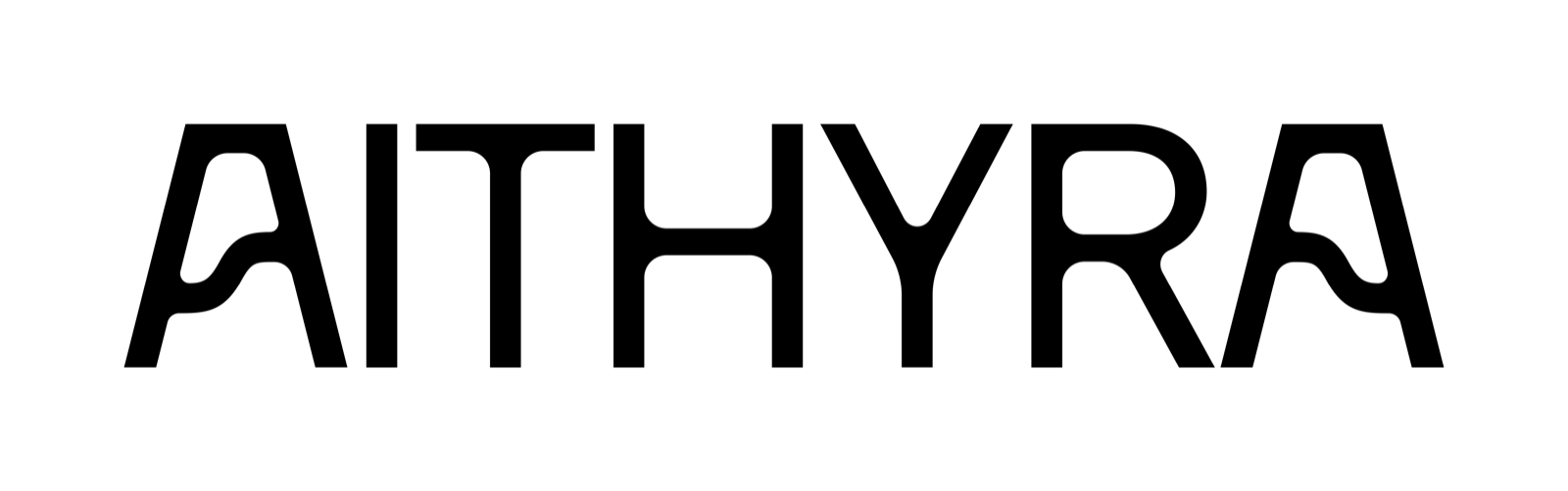}

  \noindent{\color{aithyra-teal}\rule{\textwidth}{1pt}}
  \vspace{0.05em}
  \begin{center}
    {\sffamily\bfseries\huge \@title}
  \end{center}
  \vspace{0.2em}
  \noindent{\color{aithyra-teal}\rule{\textwidth}{1pt}}
  \vspace{0.5em}
  \begin{center}
    {\sffamily\bfseries Antonio Franca\qquad Alexander Tong}\\[0.5em]
    {\sffamily\small AITHYRA, Vienna, Austria}
  \end{center}
  \vspace{0.5em}
}
\renewenvironment{abstract}{%
  \begin{tcolorbox}[
    enhanced,
    breakable,
    colback=white,
    colframe=aithyra-teal,
    boxrule=1pt,
    arc=10pt,
    left=10pt, right=10pt, top=8pt, bottom=8pt,
  ]
  {\centering\sffamily\bfseries\large\color{aithyra-teal} Abstract\par}
  \vspace{0.6em}
  \sffamily
}{%
  \par\vspace{0.8em}
  {\small\sffamily
    \noindent\textbf{Code:}
    \href{https://github.com/antoniofrancaib/dlmbench}{\texttt{https://github.com/antoniofrancaib/dlmbench}}\\
    \textbf{Correspondence:}
    \href{mailto:afranca@aithyra.ac.at}{\texttt{afranca@aithyra.ac.at}}\\
    \textbf{Date:} \@date\par
  }
  \end{tcolorbox}
}
\makeatother

\begin{document}
\maketitle

\begin{abstract}
Diffusion language models (DLMs) have emerged as the leading
non-autoregressive alternative to language modeling. 
Because their likelihoods are intractable, generative perplexity ($\genppl$) has
become the de facto standard for reporting unconditional generation quality.
We analyze why $\genppl$ is unreliable as a measure of generation quality.
Across samples, it is minimized by always emitting the scorer's most likely
sequence. Per-sample unigram entropy does not measure this loss of diversity.
Within a sample, $\genppl$ rewards predictability under the scorer, which
need not correspond to text quality.
To make this failure explicit, we construct four training-free samplers that reach or beat
trained-model $\genppl$ on LM1B and OpenWebText at non-degenerate unigram
entropy, while producing text that is incoherent by construction. 
We then
rebenchmark recent DLMs using distributional discrepancies and diversity
diagnostics, and compare their rankings with LLM judgments. As part of this
evaluation, we introduce a compression-based discrepancy that measures local
coverage and fidelity, the two directions of the KL divergence between real
and generated text. We argue that DLM progress should be evaluated with a set
of metrics that expose different properties of generated text.
\end{abstract}

% ===== HERO: Source | Generation | evaluation metrics =====
% gen-PPL, Entropy, MAUVE, GM, ED: Table~\ref{tab:rebench}, OpenWebText L=1024.
% Prefixes extended from appendix samples: reference #477, Plaid #445, Periodic #962.
% Approved restrained-color opening exhibit; held on the first page.

\definecolor{figureink}{HTML}{22282B}
\definecolor{figuremuted}{HTML}{687377}
\definecolor{figurerule}{HTML}{CED3D4}
\definecolor{figurefill}{HTML}{F3F5F5}
\definecolor{figureteal}{HTML}{0A6478}
\definecolor{figurerust}{HTML}{8D4B3C}
\newcommand{\figureheader}[1]{{\sffamily\bfseries\footnotesize #1}}
\newcommand{\figureexcerpt}[1]{{\rmfamily\fontsize{8}{10}\selectfont\raggedright\hyphenpenalty=10000\exhyphenpenalty=10000 #1\par}}
\newcommand{\figurenum}[1]{{\sffamily\fontsize{9}{10.5}\selectfont #1}}

\newcommand{\figurelabel}[2]{%
  \begin{minipage}{\linewidth}%
  \raggedright\setlength{\parindent}{0pt}\setlength{\parskip}{0pt}%
  {\sffamily\bfseries\fontsize{8.5}{10}\selectfont #1\par}%
  \vspace{1pt}%
  {\sffamily\mdseries\color{figuremuted}\fontsize{7.5}{9}\selectfont #2\par}%
  \end{minipage}%
}

\begingroup
\setlength{\intextsep}{5pt}
\begin{figure}[H]
\centering\color{figureink}
\setlength{\tabcolsep}{2.5pt}
\renewcommand{\arraystretch}{1.03}
\arrayrulecolor{figurerule}
\begin{tabularx}{\textwidth}{@{}>{\raggedright\arraybackslash}m{0.15\textwidth} X *{2}{>{\centering\arraybackslash}m{0.078\textwidth}} @ {\hspace{8pt}} *{3}{>{\centering\arraybackslash}m{0.072\textwidth}}@{}}
\toprule[0.8pt]
& & \multicolumn{2}{c}{\figureheader{gen-PPL / entropy}} & \multicolumn{3}{c}{\figureheader{Distributional}} \\
\cmidrule(lr){3-4}\cmidrule(l){5-7}
\figureheader{Source} & \figureheader{Generation} & \figureheader{gen-PPL\,$\downarrow$} & \figureheader{Entropy\,$\uparrow$} & \figureheader{MAUVE\,$\uparrow$} & \figureheader{GM\,$\downarrow$} & \figureheader{ED\,$\downarrow$} \\
\midrule[0.35pt]
\figurelabel{Reference data}{OpenWebText train} & \figureexcerpt{before I headed out I put the girls in their Vikings jammies and got ready to leave. Before I left I asked my older daughter a very important question: "Are the~\ldots} & \figurenum{\color{figuremuted} 17.1} & \figurenum{\color{figuremuted} 5.48} & \figurenum{\color{figuremuted} 1.000} & \figurenum{\color{figuremuted} 0.0} & \figurenum{\color{figuremuted} 0.00} \\[1pt]
\addlinespace[2pt]
\figurelabel{Diffusion model}{Plaid; 1.3B params} & \figureexcerpt{300 million people are infected per year. Some viruses will steal your password and make you think your password doesn't matter (e.g. Emotet). Don't click~\ldots} & \figurenum{20.7} & \figurenum{5.32} & \figurenum{0.896} & \figurenum{1.7} & \figurenum{0.43} \\[1pt]
\addlinespace[2pt]
\figurelabel{Naive generator}{Periodic-400} & \figureexcerpt{, the. to of and a in that- is for on s it with was as I: be are ( have you at by from's this he not The an has his or theyThe) said but we will " their can more who aboutt~\ldots} & \figurenum{\bfseries\color{figureteal} 19.6} & \figurenum{\bfseries\color{figureteal} 5.93} & \figurenum{\bfseries\color{figurerust} 0.004} & \figurenum{\bfseries\color{figurerust} 54.5} & \figurenum{\bfseries\color{figurerust} 28.60} \\[1pt]
\bottomrule[0.8pt]
\end{tabularx}
\arrayrulecolor{black}
\captionsetup{skip=8pt}
\caption{OpenWebText reference text, Plaid 1B, and a training-free Periodic-400 generator. Metrics: generative perplexity (with per-sample unigram entropy), MAUVE, Gradient Moment (GM), and Energy Distance (ED).}
\label{fig:hero}
\end{figure}
\endgroup

\clearpage
% =========================================================================
\section{Introduction}
\label{sec:intro}

Autoregressive (AR) models currently dominate language modeling at scale.
By factorizing the joint distribution over a token sequence into a product
of conditionals, they are trained to predict each token given its preceding
context~\citep{bengio2003neural}. In combination with the Transformer
architecture~\citep{vaswani2017attention}, this framework scales
predictably with model size and training
data~\citep{kaplan2020scaling}. However, generation remains
inherently sequential because sampling each successive token
requires conditioning on the prefix already produced. While inference
optimizations such as speculative
decoding~\citep{stern2018blockwise,leviathan2023speculative,cai2024medusa}
can reduce latency substantially, they accelerate sequential decoding
rather than decouple sampling cost from sequence length.

Diffusion language models (DLMs) take a different
paradigm based on iterative refinement~\citep{sedd2024,mdlm2024}. Rather
than committing to tokens sequentially from left to right, DLMs initialize
from a fully corrupted sequence and update multiple positions in parallel at
each step. This breaks the strict dependency between output length and
sequential computation: in principle, sequences can be generated in fewer steps than they have tokens. Although realizing competitive quality at
low step counts remains an active area of
research~\citep{arriola2025blockdiffusion,deschenaux2025sdtt}, this
computational flexibility makes DLMs an attractive direction for language
generation.

This shift in modeling paradigm, however, creates a fundamental evaluation
challenge. Progress in autoregressive modeling has long been tracked by
perplexity on fixed held-out data~\citep{lm1b2013,jozefowicz2016limits}.
When the evaluation corpus and tokenization are shared, validation
perplexity directly reflects how accurately a model captures the data
distribution, providing a reliable standard for comparison. For DLMs,
exact likelihoods under practical samplers are generally intractable.
While several model formulations admit variational or ODE-based likelihood
bounds~\citep{mdlm2024,gulrajani2023plaid,langflow2026}, the tightness of
these bounds varies across model families, complicating cross-architecture
comparisons~\citep{pynadath2026}. Consequently, the field has largely
adopted sample-based evaluation via generative perplexity
(\(\genppl\))~\citep{sedd2024,wang2025remdm}, which evaluates generated
samples under a frozen autoregressive scorer, typically
\texttt{gpt2-large}~\citep{radford2019gpt2}.

Yet \(\genppl\) does not measure generation quality in the way validation
perplexity measures model fit. Rather than assessing whether samples
resemble human text, \(\genppl\) rewards predictability under the scorer,
and it can improve substantially as sample diversity
collapses~\citep{wang2025remdm,zheng2025masked,dmmd2026}. Reporting per-sample unigram
entropy alongside \(\genppl\) has become common practice to detect
repetitive text~\citep{wang2025remdm}, but we find that this safeguard is
insufficient: even simple, incoherent samplers can achieve competitive
scores under both metrics simultaneously. In this paper, we examine why
\(\genppl\) rewards collapse, show why token-level entropy diagnostics fail
to capture mode loss, and re-evaluate recent DLMs using a broader profile of
distributional agreement, diversity, and human-aligned judgments. 

We enumerate our main contributions below:
\begin{enumerate}[leftmargin=1.4em,itemsep=2pt,parsep=0pt,topsep=2pt]
\item We analyze the information-theoretic decomposition of \(\genppl\),
showing why scoring model samples under a frozen scorer makes complete
distribution collapse mathematically optimal.

\item We construct four training-free samplers that match or outperform
trained DLMs on \(\genppl\) and unigram entropy despite generating incoherent
text.

\item We establish the limits of token-level entropy diagnostics, showing
that unigram entropy carries no information about diversity
across samples, while position-wise entropy fails to penalize cross-position dependencies.

\item We introduce Jeffreys coding divergence (JCD), a compression-based
discrepancy using context-tree weighting that measures both coverage and
fidelity without learned representations.

\item We rebenchmark recent DLMs across diverse distributional metrics and
LLM judges, showing that metric choice substantially alters model rankings
and conclusions about generation quality.
\end{enumerate}

\section{Background}
\label{sec:families}

We use the term \emph{diffusion language model} (DLM) broadly to denote any
generative model that produces token sequences by iteratively reversing a
corruption process or transporting probability mass from a base
distribution~\citep{dieleman2026continuous}. Let \(G\) denote the distribution
over complete token sequences induced by the model and a chosen sampler.
Crucially, \(G\) is determined not only by the trained model parameters, but
also by the decoding procedure~\citep{tang2026sampler}: hyperparameter choices such as the number of
sampling steps, schedule spacing, and remasking dynamics alter the output
distribution directly~\citep{peng2025pathplanning,wang2025remdm}.
Evaluating a model by standard validation perplexity would require computing
the exact likelihood \(G(x)\) on held-out sequences \(x\sim P\), where \(P\) is
the distribution of the data. As we outline below, this likelihood is
generally intractable for both discrete and continuous formulations.

\paragraph{Discrete states.}
Discrete DLMs define the generative process directly over sequence space
\(\mathcal V^L\), where \(\mathcal V\) is a discrete vocabulary and \(L\) is the
sequence length. Sampling begins from a tractable prior \(p_T\) at the maximum
noise level \(t=T\) (typically an absorbing mask state or a uniform categorical
distribution) and applies a sequence of learned reverse transition kernels
\(p_{\theta,t}(x^{(t-1)}\mid x^{(t)})\) to reach clean text at \(t=0\). The
marginal likelihood of a generated sequence \(x\) requires summing over all
valid intermediate trajectories:
\begin{equation}
G(x)
=\sum_{x^{(1)},\ldots,x^{(T)}}
p_T(x^{(T)})
\prod_{t=1}^{T}p_{\theta,t}(x^{(t-1)}\mid x^{(t)}),
\qquad x^{(0)}=x.
\label{eq:discrete-output-likelihood}
\end{equation}
Although evaluating each single-step conditional kernel is computationally
inexpensive, the number of distinct trajectories grows exponentially with
sequence length \(L\) and step count \(T\). Consequently, this marginalization
is intractable for masked diffusion and categorical state spaces more
generally~\citep{austin2021d3pm,mdlm2024,sedd2024}.

\paragraph{Continuous states.}
Continuous DLMs instead map discrete tokens into a continuous representation
space \(\mathbb R^D\) (using learned embedding spaces~\citep{li2022diffusionlm,dieleman2022cdcd},
one-hot representations~\citep{flm2026}, points on a categorical manifold~\citep{rdlm2025},
or analog bit codes~\citep{chen2023analogbits,cobit2026}) and define the diffusion or flow process over
continuous trajectories \(Z^{(t)}\). A reverse process parameterized by
\(p_{\theta,t}(z^{(t-1)}\mid z^{(t)})\) transports samples from a prior
\(Z^{(T)}\sim p_T\) (typically a standard Gaussian) to a terminal continuous state
\(Z^{(0)}\sim p_0\); in deterministic flow formulations, these reverse transitions
reduce to deterministic ODE trajectories. A decoding distribution
\(D_\phi(x\mid z)\) then maps continuous states back onto discrete sequences in
\(\mathcal V^L\):
\begin{equation}
G(x)
=\int_{\mathbb R^D}D_\phi(x\mid z)\,p_0(\mathrm dz).
\label{eq:continuous-output-likelihood}
\end{equation}
Evaluating this marginal likelihood is similarly intractable. When sampling is
stochastic, computing the terminal density \(p_0\) requires integrating over all
continuous reverse paths. Even when deterministic probability-flow ODEs render
\(p_0(z)\) tractable~\citep{song2021score}, evaluating \(G(x)\) still requires integrating over the
pre-image of continuous states that decode to \(x\). When the decoder is
deterministic (such as nearest-neighbor quantization), this integral corresponds
to computing the total probability mass of a Voronoi cell or polyhedral partition
in \(\mathbb R^D\), which cannot be calculated in closed form.

Because exact output likelihoods \(G(x)\) are intractable for both families,
evaluating DLMs requires alternative, sample-based criteria. In
Section~\ref{sec:whatmeasures}, we analyze the prevailing sample-based standard,
generative perplexity (\(\genppl\))~\citep{sedd2024,wang2025remdm}, and examine
its theoretical vulnerabilities.

% =========================================================================
\section{What generative perplexity measures}
\label{sec:whatmeasures}

Let \(X=(X_1,\ldots,X_L)\in\mathcal{V}^L\) denote a sequence of \(L\) tokens
drawn from a discrete vocabulary \(\mathcal{V}\). We denote by \(P\) the true
data distribution over \(\mathcal{V}^L\), by \(G\) the distribution over complete
samples induced by a diffusion language model and its sampler, and by \(R\)
the autoregressive scoring distribution. 

For an autoregressive model \(R\), the sequence likelihood \(R(X)\) is
directly computable via the chain rule. The population validation perplexity
\(\operatorname{ValPPL}(R)\) is defined as the exponentiated per-token
cross-entropy between the data distribution \(P\) and the model \(R\):
\begin{equation}
\operatorname{ValPPL}(R) = \exp\!\Big(\frac{1}{L}H_{\times}(P,R)\Big),
\end{equation}
where \(H_{\times}(P,R) = \mathbb{E}_{X\sim P}[-\log R(X)]\). By standard
identity, this cross-entropy decomposes into the entropy of the data and the
Kullback--Leibler (KL) divergence from the model to the data:
\begin{equation}
H_{\times}(P,R) = H(P) + D_{\mathrm{KL}}(P\|R).
\end{equation}
Because the evaluation benchmark fixes the data distribution \(P\), tokenization,
and sequence length \(L\), the data entropy \(H(P)\) is a shared constant across
all candidate models. Consequently, variations in log validation perplexity
reflect only differences in the divergence \(D_{\mathrm{KL}}(P\|R)\): a lower
score unambiguously indicates closer agreement with the target distribution.

In contrast, evaluating a DLM by validation perplexity would require computing
\(\mathbb{E}_{X\sim P}[-\log G(X)]\), which is intractable because \(G(X)\)
cannot be evaluated efficiently. Generative perplexity (\(\genppl\))
circumvents this intractability by reversing the roles of data and model: it
replaces the data distribution with the generator \(G\) and scores its samples
under a frozen autoregressive model \(R\):
\[
\mathbb{E}_{X\sim P}[-\log G(X)]
\xrightarrow{\text{reverse roles}}
\mathbb{E}_{X\sim G}[-\log P(X)]
\xrightarrow{P\to R}
\mathbb{E}_{X\sim G}[-\log R(X)].
\]
Decomposing the corresponding cross-entropy reveals a structural asymmetry
between validation and generative perplexity~\citep{pynadath2026}:
\begin{equation}
\begin{aligned}
\operatorname{ValPPL}(R)
&\mathrel{=}\exp\!\Big(\frac{1}{L}H_{\times}(P,R)\Big)\\
H_{\times}(P,R)
&=\mathbb{E}_{X\sim P}[-\log R(X)]\\
&=H(P)+D_{\mathrm{KL}}(P\|R)
\end{aligned}
\;\;\longleftrightarrow\;\;
\begin{aligned}
\operatorname{GenPPL}_R(G)
&\mathrel{=}\exp\!\Big(\frac{1}{L}H_{\times}(G,R)\Big)\\
H_{\times}(G,R)
&=\mathbb{E}_{X\sim G}[-\log R(X)]\\
&=H(G)+D_{\mathrm{KL}}(G\|R)
\end{aligned}
\label{eq:genppl}
\end{equation}
The fundamental difference lies in the leading term. Whereas \(H(P)\) is an
invariant property of the benchmark, the joint sample entropy \(H(G)\)
depends directly on the generator under test. A reduction in \(\genppl\) can
therefore arise from two conflated sources: closer distributional alignment with
the scorer (\(D_{\mathrm{KL}}(G\|R)\)), or a loss of diversity in the generated
samples (\(H(G)\)). Because \(\genppl\) measures only their sum, it cannot
distinguish realistic generation from mode collapse.

\paragraph{The metric has a collapsed optimum.}
This decomposition implies that \(\genppl\) is minimized not by matching the data
distribution, but by complete distribution collapse. Let
\(x^\star\in\arg\max_{x\in\mathcal V^L}R(x)\) denote the sequence assigned
highest probability by the scorer. For any generator \(G\),
\begin{equation}
H_{\times}(G,R)
=\sum_{x\in\mathcal V^L} G(x)[-\log R(x)]
\geq -\log R(x^\star)
=H_{\times}(\delta_{x^\star},R),
\label{eq:pointmass-optimum}
\end{equation}
with equality if and only if \(G\) places all probability mass on sequences
achieving the maximum score. Because exponentiation is strictly monotonic, the
point mass \(G=\delta_{x^\star}\) is a global minimizer of
\(\operatorname{GenPPL}_R(G)\).

By contrast, sampling from the scorer itself yields an expected cross-entropy
equal to its own entropy:
\begin{equation}
H_{\times}(R,R) = H(R) \geq -\log R(x^\star),
\end{equation}
with strict inequality whenever \(R\) is non-uniform. Consequently, \(\genppl\)
assigns a strictly better score to a collapsed sampler than to an ideal
generator that reproduces \(R\) exactly (illustrated in
Figure~\ref{fig:collapse-interpolation}, Appendix~\ref{app:proofs}).

This incentive to collapse is reinforced by the orientation of the divergence
term. In validation perplexity, \(D_{\mathrm{KL}}(P\|R) =
\mathbb{E}_{X\sim P}[\log(P(X)/R(X))]\) computes an expectation over real data.
It acts as a mean-seeking (zero-avoiding) divergence, heavily penalizing models
that fail to cover modes of \(P\). In generative perplexity, the divergence is
reversed: \(D_{\mathrm{KL}}(G\|R) = \mathbb{E}_{X\sim G}[\log(G(X)/R(X))]\)
evaluates an expectation over model generations. This objective is mode-seeking
(zero-forcing): it penalizes outputs that fall outside the high-probability
regions of \(R\), but imposes no penalty for dropping modes entirely. Thus,
both terms in Equation~\eqref{eq:genppl} favor concentrated distributions,
driving \(\genppl\) downward as diversity collapses.

\paragraph{Limits of token-level entropy diagnostics.}
To prevent trivial collapse, empirical evaluation typically reports \(\genppl\)
alongside the mean per-sample unigram entropy \(\Hemp\), defined as the entropy
of the empirical token distribution within each sample, averaged across
sample generations~\citep{zheng2025masked,wang2025remdm}. 

While this safeguard detects
repetitive outputs dominated by a single token, it is blind to collapse across
samples: a generator that deterministically outputs the exact same sample on
every draw maintains a high unigram entropy whenever that sample has a diverse
vocabulary, despite having zero sample entropy \(H(G)=0\)~\citep{acquaviva2026flowevals}. We demonstrate this
failure concretely with the training-free samplers in Section~\ref{sec:attacks}.
Correcting for \(H(G)\) directly is precluded by the fact that non-parametric
estimation of joint sequence entropy over \(\mathcal{V}^L\) from sample sizes
feasible in benchmarks is statistically impossible without strong structural
assumptions (Appendix~\ref{app:identifiability}).

An informative diagnostic is the position-wise entropy \(\Hpos\), obtained
by computing the entropy of the marginal token distribution across samples
independently at each position \(l\in\{1,\ldots,L\}\). If a generator collapses
to a single repeating sequence, every position is deterministically fixed,
yielding \(\Hpos=0\). By standard information-theoretic decomposition, the
joint entropy relates to \(\Hpos\) via:
\begin{equation}
H(G) = \sum_{l=1}^L H(X_l) - \operatorname{TC}(X_{1:L}) = L \Hpos - \operatorname{TC}(X_{1:L}),
\label{eq:joint-total-correlation}
\end{equation}
where \(\operatorname{TC}(X_{1:L}) = \sum_{l=1}^L H(X_l) - H(X_{1:L}) \geq 0\) is
the total correlation~\citep{watanabe1960information}, measuring mutual
dependence across all sequence positions. Substituting this relation into the
cross-entropy decomposition gives:
\begin{equation}
H_{\times}(G,R) - L \Hpos = D_{\mathrm{KL}}(G\|R) - \operatorname{TC}(X_{1:L}).
\label{eq:position-adjusted-decomposition}
\end{equation}
Equation~\eqref{eq:position-adjusted-decomposition} makes clear why simply
subtracting position-wise entropy fails to isolate the true model divergence
\(D_{\mathrm{KL}}(G\|R)\). The adjusted score can decrease either through
improved fidelity to the scorer or through increased total correlation across
positions, which can stem from natural linguistic structure as easily as from
rigid, repetitive patterns. We therefore treat position-wise entropy as a
diversity diagnostic rather than a basis for a corrected scalar metric. For
completeness, position-adjusted \(\genppl\) values for all benchmarked
generators are provided in Appendix~\ref{app:position-adjusted-genppl}.

% =========================================================================
\section{Constructing naive samplers with low gen-PPL}
\label{sec:attacks}

While the theoretical point-mass optimum \(G=\delta_{x^\star}\) demonstrates
that \(\genppl\) rewards collapse, it is natural to ask whether this failure
mode manifests in non-trivial generation regimes. To test whether low
\(\genppl\) can be attained without learning semantic or grammatical
structure, we construct four training-free samplers built solely from corpus
token and phrase frequencies. These samplers require no learned parameters,
optimization, or queries to the scorer. Each construction isolates and
preserves selected low-order statistics while deliberately destroying
syntactic coherence and long-range dependencies.

For the first three samplers, we let \(\hat p_k\) denote the empirical unigram
distribution over the \(k\) most frequent tokens in the training corpus,
renormalized to sum to one. We define:

\begin{itemize}[leftmargin=1.4em,itemsep=2pt,topsep=2pt]
\item \textbf{Top-\(k\) IID.}
Samples all \(L\) positions independently from \(\hat p_k\). This preserves
common unigram frequencies while completely eliminating sequential dependence
across positions (\(\operatorname{TC}(X_{1:L}) = 0\)).

\item \textbf{Mirror-\(k\).}
Samples a prefix of \(h=\lceil L/2\rceil\) tokens independently from
\(\hat p_k\), and duplicates it to form the second half of the sequence.
The marginal distribution at each individual position remains identical to that
of Top-\(k\) IID, but the exact repetition introduces a rigid inter-position
dependency that inflates total correlation \(\operatorname{TC}(X_{1:L})\) while
halving the sequence degrees of freedom.

\item \textbf{Periodic-\(k\).}
Deterministically cycles through the frequency-ordered list of the \(k\) most
common corpus tokens up to length \(L\). Because every generated sequence is
identical, the generator has zero joint entropy (\(H(G)=0\)), yet its
per-sample empirical unigram entropy \(\Hemp\) approaches \(\log k\).

\item \textbf{Phrase bank-\(m\).}
Samples 5-token n-grams uniformly at random from the \(m\) most frequent
phrases in the training corpus, concatenating them to length \(L\). While each
local block represents valid text from the data, all dependencies across phrase
boundaries are discarded.
\end{itemize}
% Zero-parameter sampler gallery (opening tokens). LM1B configurations.
\begingroup
\setlength{\intextsep}{5pt}
\begin{figure}[H]
\centering
\setlength{\fboxsep}{5pt}
\setlength{\fboxrule}{0.5pt}
\begin{minipage}[t]{0.235\linewidth}
  \fbox{\begin{minipage}[t]{\dimexpr\linewidth-2\fboxsep-2\fboxrule\relax}%
    \raggedright
    \mbox{{\footnotesize\textbf{Top-$k$ IID}}\hspace{2pt}{\fontsize{8}{9}\selectfont $k{=}64$}}\par\vspace{5pt}
    {\footnotesize\itshape said the " on on . on , to of a the of the for
    with , " is his with of has and for it\strut}
  \end{minipage}}
\end{minipage}%
\hfill%
\begin{minipage}[t]{0.235\linewidth}
  \fbox{\begin{minipage}[t]{\dimexpr\linewidth-2\fboxsep-2\fboxrule\relax}%
    \raggedright
    \mbox{{\footnotesize\textbf{Mirror}}\hspace{2pt}{\fontsize{8}{9}\selectfont $k{=}5000$}}\par\vspace{5pt}
    {\footnotesize\itshape s. for still aim Russia met Congress end Grandar
    can ,, a .Butst " paid no to resolution\strut}
  \end{minipage}}
\end{minipage}%
\hfill%
\begin{minipage}[t]{0.235\linewidth}
  \fbox{\begin{minipage}[t]{\dimexpr\linewidth-2\fboxsep-2\fboxrule\relax}%
    \raggedright
    \mbox{{\footnotesize\textbf{Periodic}}\hspace{2pt}{\fontsize{8}{9}\selectfont $k{=}400$}}\par\vspace{5pt}
    {\footnotesize\itshape the , . to of and a in '-s " that for. on is was
    withThe said as at it by from be he have has\strut}
  \end{minipage}}
\end{minipage}%
\hfill%
\begin{minipage}[t]{0.235\linewidth}
  \fbox{\begin{minipage}[t]{\dimexpr\linewidth-2\fboxsep-2\fboxrule\relax}%
    \raggedright
    \mbox{{\footnotesize\textbf{Phrase bank}}\hspace{2pt}{\fontsize{8}{9}\selectfont $m{=}5000$}}\par\vspace{5pt}
    {\footnotesize\itshape the world 's largest in the world 's , " she
    added . " the statement said . the\strut}
  \end{minipage}}
\end{minipage}%
\caption{Opening tokens of each zero-parameter sampler. Complete samples are provided in Appendix~\ref{app:gallery}.}
\label{fig:gallery}
\end{figure}
\endgroup

Figure~\ref{fig:gallery} illustrates the opening tokens produced by each
sampler. Despite their incoherence, these constructions, as we show in next section, achieve
\(\genppl\) scores competitive with, and in some cases superior to, fully
trained DLMs (Table~\ref{tab:rebench}). On LM1B, Periodic-\(64\) attains a
\(\genppl\) of \(29.4\), substantially outperforming held-out training data
(\(56.9\)) and every evaluated diffusion model. On OpenWebText,
Periodic-\(400\) achieves \(\genppl = 19.6\) with a unigram entropy
\(\Hemp = 5.93\), rivaling the top-performing diffusion models while emitting
the exact same sequence on every draw. Mirror and Phrase bank similarly reach
trained-model score ranges despite lacking grammatical syntax (parameter sweeps
across \(k\) and \(m\) are detailed in Appendix~\ref{app:sweeps}).

These constructions demonstrate that low \(\genppl\) paired with high
empirical unigram entropy does not imply sample realism or distributional
fidelity; rather, it reflects predictability under the autoregressive scorer,
which can be satisfied by matching simple marginal frequencies.

\FloatBarrier
% =========================================================================
\section{Rebenchmarking}
\label{sec:rebench}

To assess how evaluation criteria alter conclusions about generative performance,
we rebenchmark representative public DLM checkpoints on
OpenWebText~\citep{gokaslan2019openwebtext} and LM1B~\citep{lm1b2013}.
We evaluate both discrete DLMs---MDLM~\citep{mdlm2024}, SEDD~\citep{sedd2024},
Duo~\citep{duo2025}, and the SDTT teacher~\citep{deschenaux2025sdtt}---and
continuous DLMs---RDLM~\citep{rdlm2025}, Plaid~\citep{gulrajani2023plaid},
RePlaid~\citep{replaid2026}, LangFlow~\citep{langflow2026}, ELF~\citep{elf2026},
CoBit~\citep{cobit2026}, FLM, and its distilled variant FMLM~\citep{flm2026}.
For comparison, we include a standard autoregressive Transformer baseline on
OpenWebText (dataset and sampler details are documented in Appendix~\ref{app:bank}).

We evaluate generations across
three complementary axes: \textbf{distributional agreement} with the reference
corpus, \textbf{sample-level preference} from human and automated judges, and
intrinsic \textbf{sample diversity} without reference data.

\subsection{Distributional discrepancies}
\label{sec:distributional}

To evaluate distributional alignment without relying on learned representations,
we introduce Jeffreys coding divergence (JCD). Grounded in compression-based
relative entropy estimation~\citep{ziv1993relative}, JCD fits byte-level
context-tree weighting coders (CTW; \citealp{willems1995context}) to reference
and generated text. It evaluates the symmetric cross-entropy excess, measuring the
additional bits per byte required to encode held-out sequences under the opposing
model's coder. As derived in Appendix~\ref{app:ctw}, the two coding directions
provide finite-context surrogates for coverage mismatch (\(D_{\mathrm{KL}}(P\|G)\))
and fidelity mismatch (\(D_{\mathrm{KL}}(G\|P)\)).

Table~\ref{tab:rebench} summarizes the rebenchmarking results across both
corpora. The distributional metrics uniformly reject the four training-free
samplers: despite achieving competitive \(\genppl\) scores, all four samplers
receive near-zero MAUVE scores (\(\le 0.010\)) and substantially worse GM, ED,
and JCD values than every trained model.

Among trained generators, distributional rankings exhibit substantial
disagreement. No single model dominates across metrics: on both OpenWebText and
LM1B, four different generators lead at least one distributional column. On
OpenWebText, Plaid achieves the best GM (\(1.7\)) and JCD (\(0.080\)), whereas
LangFlow minimizes ED (\(0.213\)), and RePlaid achieves top scores on FBD
(\(368.8\)) and MMD (\(0.033\)). Crucially, models with low \(\genppl\)
do not necessarily achieve superior distributional agreement: CoBit-M achieves
the lowest \(\genppl\) among trained OpenWebText generators (\(19.3\)), but ranks
last among them across the distributional metrics (mean rank \(12.00\)).

% ---- REBENCHMARK TABLE (EXISTING, OWT Short + LM1B) ---------------------
\begin{table}[!ht]
\centering\fontsize{8}{9.6}\selectfont
\caption{Rebenchmarking on $1{,}024$-sample banks. Mean rank averages only the six direction-corrected distributional
ranks (MAUVE, GM, JCD, ED, FBD, MMD) among generated banks per corpus. Bold and underlining mark the best and second-best.}
\label{tab:rebench}
\setlength{\tabcolsep}{2.5pt}
\renewcommand{\arraystretch}{1.06}
\begin{tabular}{@{}lrrrrrrrrrr@{}}
\toprule
& Pref. & \multicolumn{2}{c}{Diversity} & \multicolumn{7}{c}{Distributional} \\
\cmidrule(lr){2-2}\cmidrule(lr){3-4}\cmidrule(lr){5-11}
Generator & $\genppl\downarrow$ & $\Hemp\uparrow$ & Self-BLEU$\downarrow$ & MAUVE$\uparrow$ & GM$\downarrow$ & JCD$\downarrow$ & ED$\downarrow$ & FBD$\downarrow$ & MMD$\downarrow$ & Mean rank$\downarrow$ \\
\midrule
\multicolumn{11}{@{}l}{\textbf{OpenWebText} ($L{=}1024$)}\\[1pt]
Reference (OWT train) & 17.1 & 5.48 & 0.050 & 1.000 & 0.0 & \na & 0.000 & 0.0 & 0.000 & \na \\
\addlinespace[2pt]
AR baseline & 35.3 & 5.58 & 0.047 & \best{0.925} & \second{6.1} & \second{0.113} & \third{0.411} & 372.2 & \second{0.047} & \best{2.58} \\
RePlaid 1024-NFE & 70.6 & 5.52 & 0.048 & \third{0.871} & 9.6 & 0.136 & \second{0.393} & \best{368.8} & \best{0.033} & \second{2.83} \\
Plaid 1024-NFE & \third{20.7} & 5.32 & 0.051 & \second{0.896} & \best{1.7} & \best{0.080} & 0.433 & 396.9 & 0.112 & 4.00 \\
MDLM 1024-NFE & 110.5 & \second{5.64} & \third{0.044} & 0.827 & 10.7 & 0.117 & 0.810 & 372.2 & 0.050 & 5.08 \\
LangFlow 1024-NFE & 37.0 & 5.25 & 0.049 & 0.818 & 9.9 & 0.179 & \best{0.213} & 371.9 & 0.098 & 5.17 \\
SEDD 1024-NFE & 110.0 & \third{5.62} & \best{0.037} & 0.780 & 11.0 & 0.119 & 0.844 & \third{371.8} & \third{0.049} & 5.33 \\
ELF-M 64-NFE & 22.0 & 5.20 & 0.077 & 0.862 & \third{7.6} & \third{0.115} & 0.859 & 384.0 & 0.118 & 5.67 \\
Duo 1024-NFE & 90.8 & 5.43 & \second{0.044} & 0.609 & 14.1 & 0.140 & 0.790 & \second{371.5} & 0.083 & 6.00 \\
SDTT Teacher 1024 & 54.4 & 5.31 & 0.058 & 0.587 & 12.7 & 0.159 & 1.358 & 380.2 & 0.148 & 8.33 \\
FLM 1024-NFE & 62.8 & 5.34 & 0.069 & 0.396 & 21.1 & 0.209 & 1.735 & 402.7 & 0.182 & 10.17 \\
FMLM 32-NFE & 45.1 & 5.18 & 0.137 & 0.343 & 31.4 & 0.260 & 2.473 & 409.1 & 0.257 & 11.50 \\
CoBit-M 256-NFE & \best{19.3} & 5.39 & 0.545 & 0.307 & 19.3 & 0.456 & 2.393 & 716.3 & 0.341 & 12.00 \\
\midrule
\emph{Mirror ($k{=}5000$)} & 40.8 & 5.20 & 0.053 & 0.006 & 72.9 & 0.479 & 11.191 & 707.7 & 0.214 & 13.17 \\
\emph{Periodic ($k{=}400$)} & \second{19.6} & \best{5.93} & 1.000 & 0.004 & 54.5 & 3.972 & 28.601 & 2227.6 & 0.841 & 15.08 \\
\emph{Phrase bank ($m{=}5000$)} & 60.7 & 4.63 & 0.276 & 0.010 & 116.1 & 0.884 & 8.667 & 623.6 & 0.473 & 13.50 \\
\emph{Top-$k$ IID ($k{=}64$)} & 99.9 & 3.78 & 0.144 & 0.004 & 286.8 & 4.783 & 84.266 & 1106.4 & 0.588 & 15.58 \\
\addlinespace[7pt]
\midrule[\heavyrulewidth]
\addlinespace[3pt]
\multicolumn{11}{@{}l}{\textbf{LM1B} ($L{=}128$)}\\[1pt]
Reference (LM1B train) & 56.9 & 4.33 & 0.068 & 1.000 & 0.0 & \na & 0.000 & 0.0 & 0.000 & \na \\
\addlinespace[2pt]
Duo 1024-NFE & 98.0 & 4.28 & 0.074 & \best{0.786} & \second{29.5} & \third{0.468} & \best{0.020} & \best{82.9} & 0.484 & \best{2.25} \\
MDLM 1024-NFE & 108.8 & 4.29 & 0.078 & \second{0.692} & \best{29.3} & \second{0.450} & \second{0.047} & \second{84.5} & 0.489 & \second{2.67} \\
RDLM 1000-NFE & 319.9 & \third{4.33} & \best{0.067} & 0.071 & 36.7 & \best{0.386} & \third{0.149} & 109.4 & 0.479 & 3.33 \\
FLM 1024-NFE & 97.5 & 4.28 & 0.081 & \third{0.518} & \third{32.3} & 0.481 & 0.204 & \third{89.7} & 0.484 & 3.75 \\
LangFlow 128-NFE & 139.2 & \best{4.39} & \second{0.069} & 0.009 & 52.5 & 0.510 & 1.576 & 358.0 & \best{0.368} & 5.75 \\
LangFlow 1024-NFE & 117.7 & \second{4.37} & \third{0.071} & 0.009 & 52.2 & 0.516 & 1.475 & 365.3 & \second{0.369} & 5.92 \\
FMLM 32-NFE & \third{72.1} & 4.19 & 0.101 & 0.060 & 45.5 & 0.614 & 1.859 & 118.0 & 0.521 & 6.33 \\
FMLM 4-NFE & 96.5 & 4.17 & 0.108 & 0.019 & 58.5 & 0.610 & 2.103 & 134.4 & 0.526 & 7.33 \\
FMLM 1-NFE & 115.0 & 4.13 & 0.108 & 0.007 & 82.4 & 0.681 & 3.070 & 155.7 & 0.544 & 8.83 \\
\midrule
\emph{Mirror ($k{=}5000$)} & \second{62.0} & 3.84 & 0.105 & 0.004 & 171.8 & 1.056 & 69.925 & 542.0 & \third{0.407} & 9.42 \\
\emph{Phrase bank ($m{=}1000$)} & 79.4 & 4.04 & 0.432 & 0.004 & 295.1 & 3.464 & 5.845 & 552.9 & 0.584 & 10.92 \\
\emph{Periodic ($k{=}64$)} & \best{29.4} & 4.16 & 1.000 & 0.004 & 435.0 & 5.951 & 128.144 & 900.1 & 0.964 & 12.58 \\
\emph{Top-$k$ IID ($k{=}32$)} & 75.0 & 2.99 & 0.355 & 0.004 & 462.9 & 5.288 & 42.599 & 586.3 & 0.710 & 11.92 \\
\bottomrule
\end{tabular}
\end{table}

% Final 300-rating Spearman measurements; trained-only cohort in the main text.
\begingroup
% NFE trajectories and trained-only matrix; all-bank panels are in the appendix.
% Human Elo snapshot: 2026-09-25 11:59 CEST; 300 retained ratings (304 recorded, 4 excluded).
% Sources, values, and renderer: Figures/Metric-Agreement-Human-300/.
\begin{figure}[t]
\centering
\includegraphics[width=\linewidth]{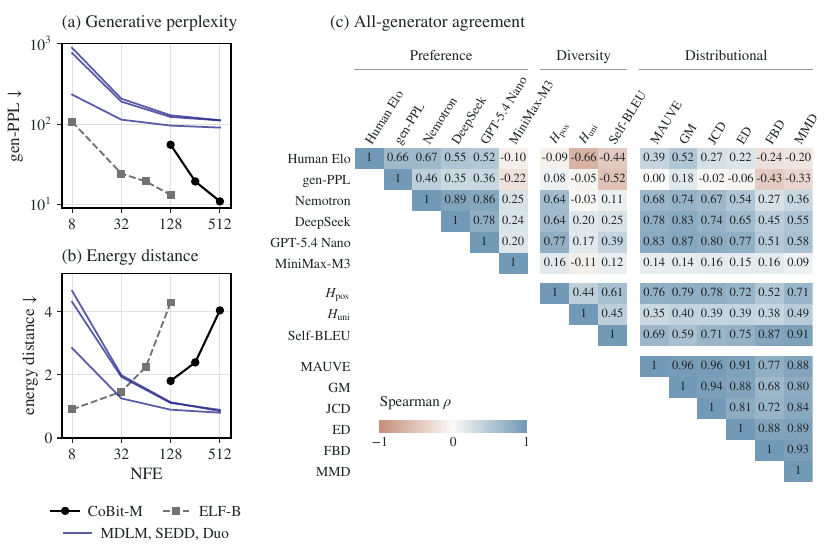}
\caption{Sampling-budget trajectories and metric agreement on OpenWebText.
\textbf{Left:} Increasing NFE lowers gen-PPL but worsens energy distance for
CoBit-M and ELF-B.
\textbf{Right:} Metric agreement over the 16 generated OpenWebText banks of
Table~\ref{tab:rebench}, excluding the real-text reference. Direction-corrected
tied-rank Spearman $\rho$; Human Elo pairs use the 12 trained generators
and the final 300-comparison snapshot.
The trained-only generators matrix appears in Appendix~\ref{app:trained-agreement}.}
\label{fig:rebenchmarking-diagnostics}
\label{fig:agreement-all}
\label{fig:nfe}
\end{figure}
\endgroup

% =========================================================================
\subsection{Preference judgments}
\label{sec:judges}

Unlike distributional discrepancies, preference metrics evaluate samples
individually. While human evaluation is the natural ground truth, comprehensive
human studies across models and sampling budgets are cost-prohibitive. We
therefore collect blinded pairwise judgments from four language models of
varying scale---Nemotron-3-Ultra-550B-A55B, DeepSeek, GPT-5.4 Nano, and
MiniMax-M3---and convert their win rates into standard Elo
ratings~\citep{elo1978rating}.

As reported in Table~\ref{tab:judges}, three of the four LLM judges place every
naive sampler strictly below all trained DLMs, rejecting the artificial
predictability that optimizes \(\genppl\). For example, Periodic-\(400\), which
achieves the second-best \(\genppl\) on OpenWebText, ranks last under Nemotron
Elo (\(679\)). The only exception is MiniMax-M3, which frequently designates
both samples in a pair as poor, leading to a high rate of ties that diminishes
discriminative power.

To calibrate these automated judgments against human perception, we conducted a
controlled human evaluation on the twelve trained OpenWebText generators and real
text. Nine human evaluators reviewed 35-word prefixes from anonymized pairs,
selecting the sample with superior fluency, coherence, and local structure
(or skipping if indecisive). We fit Elo ratings to the 300 retained
comparisons using the same procedure applied to the LLM judges.

Given the sample size (several dozen comparisons per generator), we interpret
the human ratings in terms of broad tiers rather than fine-grained rank
orderings. Within this scope, Nemotron Elo exhibits meaningful rank agreement
with human evaluators (\(\rho = 0.67\); Figure~\ref{fig:agreement-all}, right),
with both human raters and Nemotron identifying Plaid and ELF-M as the top
generative models. Conversely, MiniMax-M3 shows negligible correlation with human
judgments, consistent with its high tie rate. These findings support using
suitably scaled LLM judges (such as Nemotron) as sample-level quality proxies,
provided they are paired with diversity diagnostics.

\begin{table}[!htb]
% Approved Elo tables. Black text explicitly requested by Antonio on 2026-09-24.
% Human Elo: final 300-comparison snapshot retrieved 2026-09-25 11:59 CEST.
% Data: Figures/Metric-Agreement-Human-300/human_snapshot.json.
% Nemotron ratings retain the full 21-bank campaign; rows remain in Nemotron order.
% Match counts and cohort policies are documented in Appendix E.
% Source: Lab/Pages/scalable-llm-judges/Views/result-table-ca316a8bd1e6/
% Data/865e1449ed6f-pairwise.json; rows remain in Nemotron order.
% Ratings retain the full 21-bank OWT cohort.
% Human snapshot: Figures/Metric-Agreement-Human-300/human_snapshot.json (300 comparisons).
\centering\footnotesize
\caption{Pairwise Elo on OWT from four LLM judges and humans
(higher is better; best generated scores in bold, second best underlined).
Human ratings use 300 comparisons of 35-word prefixes from 12 trained
generators and real text. Elo scales differ across judges.}
\label{tab:judges}
\label{tab:human-elo}
\label{tab:judges-other-owt}
\renewcommand{\arraystretch}{1.04}
\setlength{\tabcolsep}{3pt}
\begin{tabular*}{\linewidth}{@{\extracolsep{\fill}}lrrrrr@{}}
\toprule
Generator & Nemotron & DeepSeek & GPT-5.4 Nano & MiniMax M3 & Human \\
\midrule
Reference (OWT train) & 2553 & 2091 & 1835 & 1699 & 1554.9 \\
\addlinespace[2pt]
Plaid 1024-NFE & \textbf{2026} & \textbf{1761} & \textbf{1767} & \textbf{1532} & \underline{1593.8} \\
ELF-M 64-NFE & \underline{1817} & 1536 & \underline{1667} & 1488 & \textbf{1628.0} \\
CoBit-M 256-NFE & 1691 & 1527 & 1516 & 1489 & 1458.6 \\
AR baseline & 1681 & \underline{1555} & 1626 & 1478 & 1510.9 \\
FMLM 32-NFE & 1565 & 1513 & 1469 & 1496 & 1469.5 \\
SDTT Teacher 1024 & 1548 & 1514 & 1451 & 1490 & 1528.2 \\
LangFlow 1024-NFE & 1527 & 1511 & 1564 & 1488 & 1552.3 \\
RePlaid 1024-NFE & 1513 & 1540 & 1481 & 1492 & 1495.4 \\
MDLM 1024-NFE & 1505 & 1450 & 1489 & 1489 & 1405.7 \\
FLM 1024-NFE & 1495 & 1492 & 1429 & 1488 & 1502.2 \\
SEDD 1024-NFE & 1477 & 1513 & 1451 & 1488 & 1349.2 \\
Duo 1024-NFE & 1447 & 1472 & 1491 & 1502 & 1451.3 \\
\midrule
\emph{Phrase bank ($m{=}5000$)} & 1294 & 1343 & 1404 & 1469 & \na \\
\emph{Mirror ($k{=}5000$)} & 727 & 1220 & 1337 & 1482 & \na \\
\emph{Top-$k$ IID ($k{=}64$)} & 709 & 1258 & 1216 & \underline{1503} & \na \\
\emph{Periodic ($k{=}400$)} & 679 & 1271 & 1279 & 1483 & \na \\
\bottomrule
\end{tabular*}
\end{table}

\subsection{Diversity diagnostics}
\label{sec:diversity}

Because sample-level preference evaluators score outputs in isolation, they are
fundamentally incapable of assessing distribution support or penalizing mode
collapse; in this respect, LLM judges share the failure mode of \(\genppl\)
and human evaluation~\citep{huse2019}. Consequently, preference
scores must be interpreted alongside explicit diversity diagnostics.

Figure~\ref{fig:rebenchmarking-diagnostics} (left) illustrates the operational
necessity of this distinction across sampling budgets. As the number of
function evaluations (NFE) increases, both CoBit-M and ELF-B achieve lower
\(\genppl\) than real human text, yet their energy distance to the data
distribution steadily deteriorates. The diversity diagnostics reveal two
distinct mechanisms driving this divergence:
\begin{itemize}[leftmargin=1.4em,itemsep=2pt,topsep=2pt]
\item For CoBit-M, samples become increasingly homogeneous across draws,
reflected in a sharp rise in Self-BLEU (reaching \(0.545\) at 256 NFE).
\item For ELF-B, samples exhibit increasing internal token repetition,
reflected in a progressive drop in empirical unigram entropy \(\Hemp\).
\end{itemize}
In both cases, additional sampling steps concentrate probability mass on regions
that the autoregressive scorer finds predictable, at the expense of modeling the
full data distribution.

Diversity is also where preference signals and distributional metrics diverge
empirically. Whereas distributional discrepancies correlate positively with both
diversity diagnostics (Figure~\ref{fig:rebenchmarking-diagnostics}, right),
preference signals exhibit weak or negative correlations with sample variety;
in particular, \(\genppl\) tends to improve as sample repetition intensifies.
Moreover, no single diversity diagnostic is sufficient: Periodic-\(400\)
achieves the highest unigram entropy in the benchmark (\(\Hemp = 5.93\)) despite
complete collapse (\(\text{Self-BLEU} = 1.000\)), while Mirror-\(5000\) evades
both by repeating blocks within samples. Robust evaluation requires combining
within-sample, position-wise, and distributional measurements.

% =========================================================================
\section{Evaluation as a metric profile}
\label{sec:claims}

% Human correlations below use the final 300-rating, 12-trained-generator snapshot:
% Figures/Metric-Agreement-Human-300/spearman-trained-correlations.csv.
% Supporting calculations: Figures/Metric-Profile-Evidence/metric_profile_evidence.json.
% Its human-dependent results use the same final snapshot; automatic measurements
% retain their frozen metric, position-entropy, and encoder-sensitivity inputs.
Our findings indicate that unconditional text evaluation cannot be reliably
captured by a single scalar objective. Instead, we recommend evaluating
generators through a metric profile combining sample-level preference signals
with distributional discrepancies and explicit diversity guardrails. Because
these measurements assess distinct statistical properties of generated text,
reporting them jointly exposes pathological behaviors that any single evaluator
overlooks.

\paragraph{Vulnerability of sample-level preference to collapse.}
Any metric that scores sequences independently is susceptible to
mode collapse. For any sample-level score \(s(x)\) (or expected pairwise outcome
against fixed opponents), linearity of expectation gives
\begin{equation}
\mathbb{E}_{X\sim G}[s(X)] = \sum_{x\in\mathcal{V}^L} G(x) s(x) \le \max_{x\in\mathcal{V}^L} s(x),
\end{equation}
with equality achieved when \(G\) collapses to a point mass on any maximizing
sequence. Equation~\eqref{eq:pointmass-optimum} is the special case \(s(x)=\log R(x)\),
but this applies to any samplewise judge: in automated evaluations, Nemotron Elo
systematically increases as generator diversity declines.
Preference metrics must therefore be paired with multi-scale entropy guardrails:
within-sample variety \(\Hemp\) to detect internal repetition, and position-wise
entropy \(\Hpos\) to detect cross-sample collapse. Because \(\Hpos = 0\) under
complete collapse, it catches failures like Periodic-\(400\) that retain high
\(\Hemp\), while \(L\Hpos\) provides a direct, estimable upper bound on joint
entropy \(H(G)\) (Equation~\eqref{eq:joint-total-correlation}; comparisons require
matched bank sizes).

\paragraph{Representation dependence in distributional discrepancies.}
Distributional discrepancies compare entire sample collections, avoiding the
linearity of samplewise metrics~\citep{he2023blindspots}. By the data-processing
inequality for \(f\)-divergences~\citep{sason2019dataprocessing},
\(D(P_T\|G_T)\le D(P\|G)\) for any representation \(T\); feature-space distances
merely lower-bound true sequence divergence. Crucially, representation choice
dominates metric choice: under a fixed encoder, MAUVE, Fr\'echet distance,
Energy Distance, and MMD produce nearly identical rankings, whereas altering the
encoder shifts models like ELF-M from second to ninth place
(Appendix~\ref{app:encoder-sensitivity}). A robust profile must therefore combine
discrepancies built on fundamentally different representations.

\paragraph{LLM judges versus generative perplexity.}
For sample quality, we favor calibrated LLM-judge Elo over \(\genppl\).
Generative perplexity relies on a double proxy: an AR model substitutes for a
human, and likelihood substitutes for quality. LLM judges eliminate the second
proxy by evaluating quality directly. In our experiments, three of four LLM
judges correlate positively with distributional metrics, whereas \(\genppl\)
tracks neither distributional agreement nor diversity
(Figure~\ref{fig:agreement-all}, right). % Nemotron Elo also aligns with human evaluations (\(\rho = 0.67\)), provided prompt and protocol are explicitly reported.

\paragraph{Recommended metric profile.}
We recommend evaluating unconditional generation through a two-part profile:
(1)~\textbf{Sample quality with diversity guardrails:} pairwise LLM-judge Elo
(such as Nemotron), reported alongside within-sample \(\Hemp\) and position-wise
\(\Hpos\); and (2)~\textbf{Distributional agreement across representations:}
Gradient Moment (parameter gradients), MAUVE (GPT-2 features), and Energy
Distance (geometric features), supplemented by encoder-free Jeffreys coding
divergence (JCD). Divergence across these axes signals collapse, as seen along
the NFE trajectories of CoBit-M and ELF-B
(Figure~\ref{fig:rebenchmarking-diagnostics}, left). Where historical benchmarks
require \(\genppl\), dividing by \(\exp(\Hpos)\) removes the reward for
cross-sample collapse (Appendix~\ref{app:position-adjusted-genppl}).

% =========================================================================
\section{Conclusion}
\label{sec:conclusion}

Generative perplexity evaluates scorer predictability rather than faithful text
modeling. Because scoring samples places generator joint entropy into the
objective, \(\genppl\) is minimized by distribution collapse, which unigram
entropy cannot detect. Zero-parameter samplers outperform trained models on this
metric, and higher sampling budgets can lower \(\genppl\) while moving generations
away from data. We recommend evaluating DLMs through the profile of
Section~\ref{sec:claims} across sampling budgets (NFE), documenting
precision~\citep{zheng2025masked}, and verifying that proposed metrics rank naive
heuristics last. While our human study evaluates short prefixes, automated
judges may carry subtle biases~\citep{wataoka2024selfpreference}, and token
guardrails miss structured dependencies like Mirror, balanced metric profiles
provide a necessary foundation for evaluating genuine progress in
non-autoregressive language generation.

\subsection*{Acknowledgments}
We thank Raul Minan, Maximilian Guidi, Zander Blasingame, Luka Mucko, Sebastian
Bauer, Angelo Gnazzo, and Katarina Petrovic for human preference evaluations. We
also thank Raul Minan, Niklas Rindtorff, Fred Peng, Shucheng Li, and Luka Mucko
for valuable feedback on the manuscript and ideas. This research was enabled by
computational resources provided by AITHYRA, supported by the Austrian Academy of
Sciences and the not-for-profit Boehringer Ingelheim Stiftung.

\clearpage
\newrefcontext[sorting=nyt]
\printbibliography[heading=bibintoc]

% =========================================================================
\appendix
% Auto-generated appendix contents: lists the sections and subsections
% below, and updates itself as the appendix changes.
\newpage
\crefalias{section}{appendix}
\crefalias{subsection}{appendix}
\crefalias{subsubsection}{appendix}
\startcontents[]
\printcontents[]{l}{1}[2]{{\bfseries \large Appendix contents}}
\vspace{10pt}

\clearpage
\renewcommand{\topfraction}{0.92}
\renewcommand{\bottomfraction}{0.80}
\renewcommand{\textfraction}{0.08}
\renewcommand{\floatpagefraction}{0.80}
\setcounter{topnumber}{3}
\setcounter{bottomnumber}{2}
\setcounter{totalnumber}{4}
\setlength{\floatsep}{10pt plus 2pt minus 2pt}
\setlength{\textfloatsep}{12pt plus 2pt minus 2pt}
\setlength{\intextsep}{10pt plus 2pt minus 2pt}
\captionsetup[figure]{skip=6pt,belowskip=0pt}
\captionsetup[table]{skip=6pt,belowskip=0pt}
\titlespacing*{\section}{0pt}{15pt plus 3pt minus 2pt}{7pt plus 1pt minus 1pt}
\titlespacing*{\subsection}{0pt}{12pt plus 2pt minus 2pt}{5pt plus 1pt minus 1pt}
\titlespacing*{\subsubsection}{0pt}{10pt plus 2pt minus 1pt}{4pt plus 1pt minus 1pt}
\makeatletter
\setlength{\@fptop}{0pt}
\setlength{\@fpsep}{12pt}
\makeatother

\section{Entropy and the gen-PPL decomposition}
\label{app:entropy}

In this appendix, we formalize the four entropy quantities used throughout
the paper and prove their ordering. We then examine why directly subtracting
an entropy estimate from \(\genppl\) fails to recover a sound evaluation
metric without strong structural assumptions, and analyze why position-wise
entropy adjustments remain vulnerable to structured repetition.

\subsection{Entropy quantities and their ordering}
\label{app:proofs}

Let \(X = (X_1, \ldots, X_L) \sim G\) denote a sequence of length \(L\) drawn
from generator \(G\). For any token \(v \in \mathcal{V}\), let
\(p_t(v) = \Pr_G(X_t = v)\) be the marginal distribution at position \(t\),
\(\bar{p}(v) = \frac{1}{L} \sum_{t=1}^L p_t(v)\) the marginal pooled across all
positions, and \(\widehat{p}_x(v) = \frac{1}{L} \sum_{t=1}^L \mathbf{1}[x_t = v]\)
the empirical unigram histogram of an individual sample \(x\). These definitions
give rise to four distinct entropy measures:
\begin{align}
H_{\mathrm{pool}} &= H(\bar{p}), &
H_{\mathrm{uni}} &= \mathbb{E}_{X \sim G}\bigl[H(\widehat{p}_X)\bigr], \nonumber\\
H_{\mathrm{pos}} &= \frac{1}{L} \sum_{t=1}^{L} H(p_t), &
H_{\mathrm{joint}} &= H(G) = H(X_1, \ldots, X_L).
\label{eq:four-entropies}
\end{align}
The first three quantities are measured in nats per token. To compare them on a
shared scale, we normalize the joint sequence entropy by sequence length,
\(H_{\mathrm{joint}} / L\).

Each quantity discards different structural information:
\begin{itemize}[leftmargin=1.4em,itemsep=2pt,parsep=0pt,topsep=2pt]
\item Pooled entropy \(H_{\mathrm{pool}}\) reflects aggregate vocabulary use
across the entire generated corpus, ignoring both position and sample boundaries.
\item Within-sample unigram entropy \(H_{\mathrm{uni}}\) captures lexical
diversity inside individual sequences, but discards token order and cross-sample
variation entirely.
\item Position-wise entropy \(H_{\mathrm{pos}}\) captures uncertainty across
samples at each coordinate, but treats positions independently.
\item Only the joint sequence entropy \(H_{\mathrm{joint}}\) measures true
uncertainty over complete generated sequences.
\end{itemize}

Concavity of the entropy functional implies that pooling distributions cannot
decrease entropy, yielding \(H_{\mathrm{uni}} \le H_{\mathrm{pool}}\) and
\(H_{\mathrm{pos}} \le H_{\mathrm{pool}}\), because \(\bar{p}\) is the mixture
of individual sample histograms as well as position marginals.
Subadditivity of joint entropy ensures that \(H_{\mathrm{joint}} / L \le H_{\mathrm{pos}}\),
where the gap is exactly the normalized total correlation~\citep{watanabe1960information}:
\begin{equation}
L\,H_{\mathrm{pos}} - H(G)
= \operatorname{TC}(X_{1:L})
= D_{\mathrm{KL}}\Bigl(G \,\Big\|\, \prod_{t=1}^L p_t\Bigr) \ge 0,
\label{eq:app-total-correlation}
\end{equation}
which corresponds to the identity in Equation~\eqref{eq:joint-total-correlation}.
By contrast, \(H_{\mathrm{uni}}\) and \(H_{\mathrm{pos}}\) cannot be ordered in
general: neither bounds the other, and neither suffices to determine
\(H_{\mathrm{joint}}\).

Our synthetic generators illustrate both failure modes directly:
\begin{itemize}[leftmargin=1.4em,itemsep=2pt,parsep=0pt,topsep=2pt]
\item \textbf{Periodic-400} repeatedly emits the same token-rich sequence.
Because each sample contains hundreds of distinct words, it achieves high
within-sample diversity (\(H_{\mathrm{uni}} \approx H_{\mathrm{pool}} \approx 5.93\)).
Yet because outputs never vary across runs, its cross-sample uncertainty is
zero: \(H_{\mathrm{pos}} = H_{\mathrm{joint}} = 0\).
\item \textbf{Mirror} matches the exact position marginals of Top-\(k\) IID at
every token, yielding identical \(H_{\mathrm{pos}}\). However, because it
deterministically copies its first half into its second, it doubles positional
dependencies and cuts the true joint entropy in half:
\(H(G) = \frac{1}{2} H_{\mathrm{top-}k}\).
\end{itemize}

\begin{figure}[!htbp]
\centering
\includegraphics[width=0.76\linewidth]{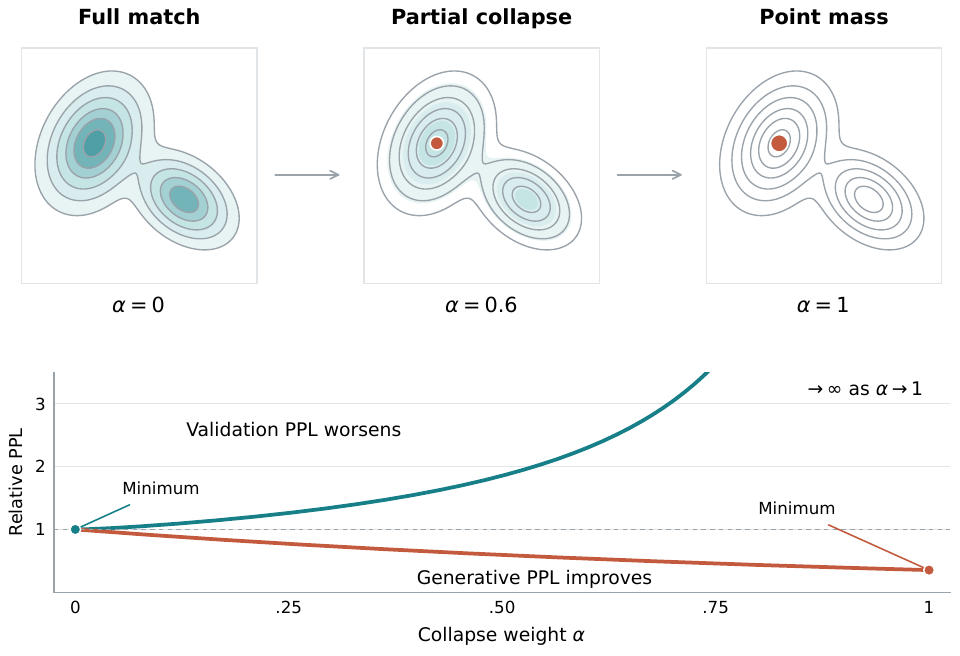}
\caption{Interpolation from distribution matching (\(\alpha=0\)) to collapse
(\(\alpha=1\)), along the path
\(G_\alpha=(1-\alpha)R+\alpha\delta_{x^\star}\). Gray contours show \(R\) and
color shows \(G_\alpha\).}
\label{fig:collapse-interpolation}
\end{figure}

Figure~\ref{fig:collapse-interpolation} illustrates the collapsed optimum
derived in Section~\ref{sec:whatmeasures} along the mixture path
\(G_\alpha = (1-\alpha)R + \alpha\delta_{x^\star}\), where \(x^\star = \arg\max_x R(x)\)
is the scorer's most probable sequence. Because cross-entropy is strictly linear
in the mixture parameter \(\alpha\),
\[
H_{\times}(G_\alpha, R) = (1-\alpha)\,H(R) - \alpha\log R(x^\star),
\]
and because \(-\log R(x^\star) < H(R)\) for any non-degenerate distribution, the
score decreases monotonically as probability shifts from distribution matching
(\(\alpha=0\)) toward complete collapse (\(\alpha=1\)). In other words, every step
toward the point mass improves \(\genppl\).

\subsection{Why entropy subtraction requires assumptions}
\label{app:identifiability}

At first glance, one might hope to repair \(\genppl\) simply by subtracting the
generator's joint entropy \(H(G)\), leaving
\(H_\times(G,R) - H(G) = D_{\mathrm{KL}}(G\|R)\). Even in this idealized scenario,
however, two fundamental obstacles remain:

First, even assuming an ideal scorer (\(R = P\)), the resulting difference
yields \(D_{\mathrm{KL}}(G\|P)\). This is the mode-seeking direction of the KL
divergence: it heavily penalizes a generator for placing probability mass where
real text is rare, but assesses zero penalty for dropping entire modes of the
data distribution. It remains fundamentally different from the coverage-sensitive
divergence \(D_{\mathrm{KL}}(P\|G)\) that standard validation perplexity targets.

Second, in practice \(H(G)\) is unknown and cannot be estimated reliably from
benchmark sample banks without severe structural assumptions. For sequences of
length \(L=1024\) over a vocabulary of size \(V \approx 50{,}000\), the discrete
support size \(V^L\) dwarfs any practical evaluation budget. On a typical
benchmark bank of \(N = 1{,}024\) samples, virtually every generated sequence is
unique. A naive plug-in estimator simply returns the empirical upper bound
\(\log N\), regardless of whether the model's true support spans hundreds or
billions of sequences. Minimax lower bounds for discrete entropy estimation
formalize this limitation: unbiased estimation without prior structural
constraints is statistically impossible unless sample size scales with the
support cardinality~\citep{valiant2011estimating,wu2016minimax}. Reliable
entropy estimation is possible only by imposing explicit constraints, such as
finite-order Markov chains or context-tree structures.

\subsection{Position-wise entropy adjustment}
\label{app:position-adjusted-genppl}

While estimating joint sequence entropy is intractable, its upper bound—the
position-wise entropy \(\Hpos\)—can be computed reliably from sample banks
because it requires only marginal token frequencies at each coordinate \(t\).
This motivates testing an adjusted score that divides \(\genppl\) by \(\exp(\Hpos)\).

\begin{figure}[!htbp]
\centering
\includegraphics[trim=22bp 30bp 14bp 22bp,clip,width=0.66\linewidth]{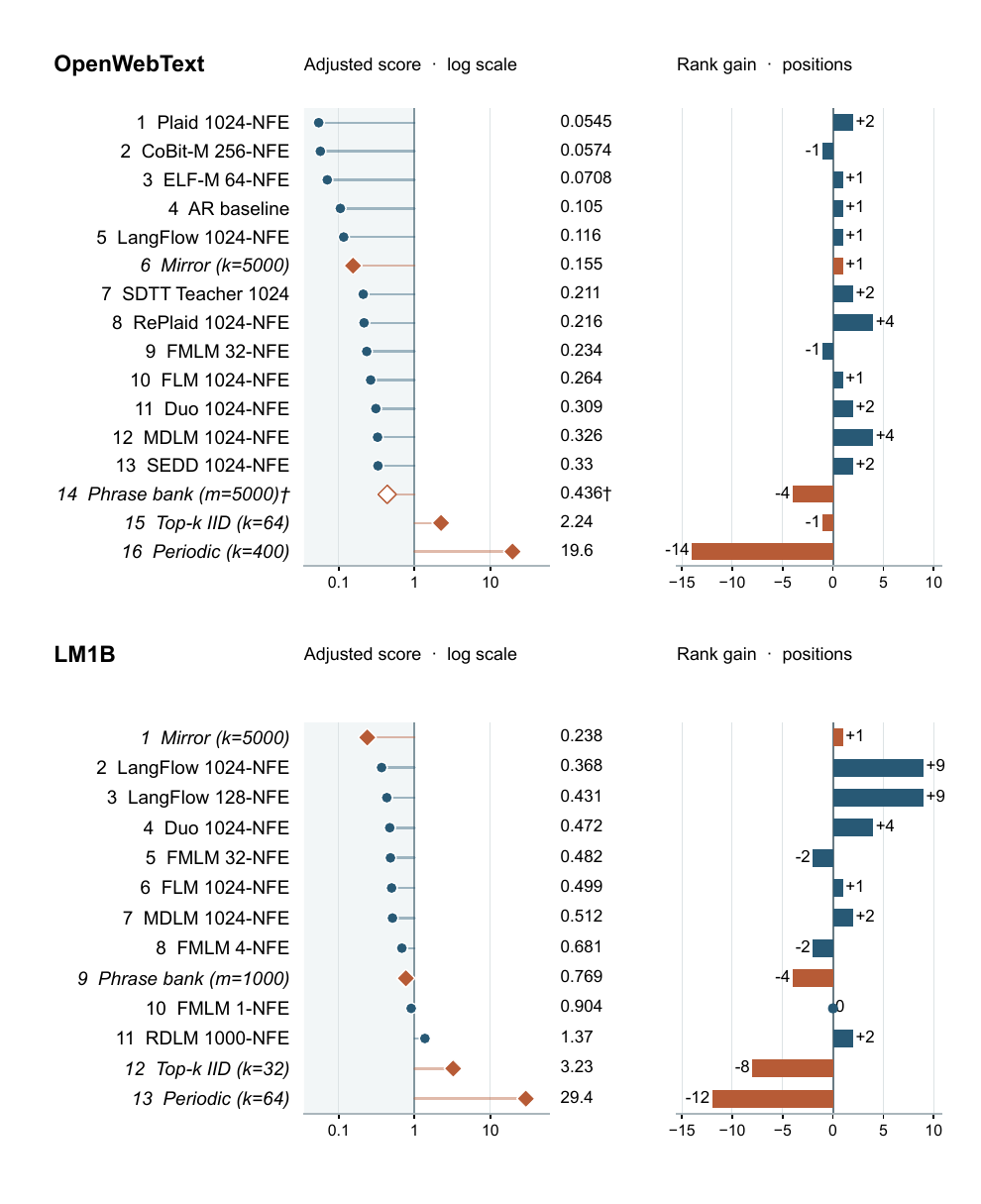}
\caption{Position-wise entropy adjustment for the generators in
Table~\ref{tab:rebench}. Left: \(\genppl\) divided by \(\exp(\Hpos)\), log
scale. Right: original \(\genppl\) rank minus adjusted rank.}
\label{fig:position-adjusted-genppl}
\end{figure}

Substituting the total correlation identity from
Equation~\eqref{eq:joint-total-correlation} reveals what this adjustment
actually computes:
\begin{equation}
\frac{\genppl(G; R)}{\exp(\Hpos)}
= \exp\left( \frac{1}{L} \Bigl( D_{\mathrm{KL}}(G\|R) - \operatorname{TC}(X_{1:L}) \Bigr) \right).
\label{eq:app-adjusted-genppl}
\end{equation}
As shown in Figure~\ref{fig:position-adjusted-genppl}, this adjustment successfully
eliminates the artificial advantage of repetition across runs. For Periodic,
\(\Hpos = 0\), leaving its score unchanged, whereas genuinely diverse generators
see their adjusted scores substantially reduced.

However, the adjusted metric still holds a failure mode. Because
Equation~\eqref{eq:app-adjusted-genppl} subtracts the total correlation
\(\operatorname{TC}(X_{1:L})\), a generator can improve its score simply by
introducing spurious dependencies across positions. For example, our naive sampler Mirror
copies its first 512 tokens directly into its second 512 tokens. This deterministic
repetition inflates total correlation, allowing Mirror to achieve a better adjusted
score than any trained diffusion model in our benchmark. Thus, while position-wise
adjustment penalizes cross-sample collapse, it remains fundamentally vulnerable to
structured repetition.

% =========================================================================
\section{Benchmark protocol}
\label{app:rebenchmark-details}

To ensure fair and reproducible comparisons across model families, our
benchmark protocol holds the evaluation environment strictly constant.
Evaluating non-autoregressive language models can easily be confounded by
differences in decoding schedules, reference splits, and metric
implementations. We eliminate these discrepancies by evaluating all models on
identical, pre-generated sample banks using frozen metric pipelines.

\subsection{Frozen metric configuration}
\label{app:frozen}

Comparing distributional metrics across generators requires fixing both the
reference corpus and the estimator hyperparameters. For each dataset, we draw a
single reference bank of \(1{,}024\) data samples and evaluate all
generators against it. All metrics operate on these exact sample banks with
frozen configurations (summarized in Table~\ref{tab:frozen-config}).

For JCD, which fits separate context trees to real and generated text, we
partition the \(1{,}024\) samples into \(512\) documents for fitting the coders
and \(512\) held-out documents for cross-evaluation; all other metrics score the
full \(1{,}024\) samples. Sample lengths are capped at the native corpus
limits (\(128\) tokens for LM1B and \(1{,}024\) tokens for OpenWebText) without
adding artificial sequence delimiters.

\begin{table}[!htbp]
\centering\footnotesize
\caption{Frozen configuration of the evaluation metrics in
Table~\ref{tab:rebench}. Max tokens equals the corpus sequence length
(\(128\) for LM1B, \(1{,}024\) for OpenWebText) unless stated.}
\label{tab:frozen-config}
\setlength{\tabcolsep}{5pt}
\begin{tabular}{@{}lp{0.72\linewidth}@{}}
\toprule
Metric & Configuration \\
\midrule
gen-PPL, \(\Hemp\) & \texttt{gpt2-large} scorer; \(\exp\) of the mean per-sample
next-token NLL; batch 8; text input without added special tokens. \\
MAUVE & \texttt{mauve-text} 0.4.0; \texttt{gpt2-large} features, 512 tokens;
\(k\)-means on features explaining 90\% variance (5 restarts, 500 iterations);
scaling factor 5; five subset seeds at \(n{=}1{,}024\) per side. \\
FBD & Fr\'echet distance between Gaussian fits of \texttt{gpt2-large}
features. \\
GM & Squared difference of mean parameter gradients under \texttt{gpt2};
batch 4. \\
ED & Energy distance on handcrafted entity and discourse features,
standardized by reference mean and scale. \\
MMD & Unbiased \(\mathrm{MMD}^2\) with an L2-normalized character
3--5-gram cosine kernel; \(1{,}024\) balanced rows, selection seed 1729. \\
JCD & CTW-20 over UTF-8 bits; 512 fitting and 512 held-out documents per
source; split seeds 0--4 (Appendix~\ref{app:ctw}). \\
Self-BLEU & Mean pairwise within-bank sentence BLEU~\citep{papineni2002bleu}, \(n\)-gram orders
2--4. \\
\bottomrule
\end{tabular}
\end{table}

\subsection{Sample banks and checkpoint sources}
\label{app:bank}

We evaluate public checkpoints using the authors' official implementations and
recommended default sampling parameters, avoiding post-hoc tuning. Each
evaluated model produces a bank of \(1{,}024\) samples generated under a fixed
seed schedule. To guarantee reproducibility, each bank is accompanied by an
immutable manifest recording the exact checkpoint digest, code commit, and job
environment. The sole exception is RePlaid, for which pre-generated sample banks
were provided directly by the authors~\citep{replaid2026}.
Table~\ref{tab:checkpoint-sources} lists the checkpoint sources for all trained
generators in Table~\ref{tab:rebench}, and Appendix~\ref{app:gallery} provides
qualitative sample galleries.

\begin{table}[!htbp]
\centering\scriptsize
\caption{Sources of the trained generators in Table~\ref{tab:rebench}.
Hugging Face (HF) and GitHub (GH) locations are those from which checkpoint
weights were obtained.}
\label{tab:checkpoint-sources}
\setlength{\tabcolsep}{3pt}
\begin{tabular*}{\linewidth}{@{\extracolsep{\fill}}lll@{}}
\toprule
Generator & Corpus & Checkpoint source \\
\midrule
AR baseline & OWT & GH \texttt{s-sahoo/duo}~\citep{duo2025} \\
MDLM & OWT & HF \texttt{kuleshov-group/mdlm-owt}~\citep{mdlm2024} \\
MDLM & LM1B & GH \texttt{s-sahoo/duo}~\citep{duo2025} \\
SEDD & OWT & HF \texttt{louaaron/sedd-medium}~\citep{sedd2024} \\
Duo & OWT, LM1B & GH \texttt{s-sahoo/duo}~\citep{duo2025} \\
SDTT Teacher & OWT & HF \texttt{jdeschena/sdtt} (\texttt{teacher\_1M\_sm})~\citep{deschenaux2025sdtt} \\
FLM, FMLM & OWT, LM1B & GH \texttt{david3684/flm}~\citep{flm2026} \\
LangFlow & OWT, LM1B & HF \texttt{Continuous-Rivals-Discrete}~\citep{langflow2026} \\
RDLM & LM1B & GH \texttt{harryjo97/RDLM}~\citep{rdlm2025} \\
ELF-M & OWT & HF \texttt{embedded-language-flows/ELF-M-owt-torch}~\citep{elf2026} \\
CoBit-M & OWT & HF \texttt{gbatzolis/CoBit} (EMA, 750k steps)~\citep{cobit2026} \\
Plaid 1B & OWT & GH \texttt{igul222/plaid}~\citep{gulrajani2023plaid} \\
RePlaid & OWT & Samples provided by the authors~\citep{replaid2026} \\
\bottomrule
\end{tabular*}
\end{table}

\subsection{Attack parameter sweeps}
\label{app:sweeps}

A natural question is whether the competitive scores of our training-free
samplers depend on fortunate hyperparameter choices. To address this, we sweep
the primary parameter governing each family: vocabulary cutoff \(k\) for
Top-\(k\) IID, Mirror-\(k\), and Periodic-\(k\), and bank size \(m\) for Phrase
bank-\(m\). Figure~\ref{fig:frontier} plots the resulting frontiers in the
\(\genppl\)--\(H_{\mathrm{uni}}\) plane against the range spanned by trained
models.

These sweeps demonstrate that the vulnerability of \(\genppl\) is broad and
systematic. On OpenWebText, Periodic configurations across an entire range of
\(k\) achieve \(\genppl\) between \(4.73\) and \(7.14\)—substantially
outperforming real human text (\(17.1\))—while maintaining unigram entropies
(\(5.18\)--\(5.64\) nats) directly inside the range of trained generators. The
configurations in Table~\ref{tab:rebench} (such as Periodic-400 at \(\genppl =
19.6\) and unigram entropy \(5.93\)) were fixed prior to the rebenchmark, and
neighboring settings yield similarly pathological scores. This confirms that
scoring samples under a frozen autoregressive likelihood creates an extensive
region of degenerate optima that simple heuristics readily populate.
\begin{figure}[!htbp]
\centering
\begin{tikzpicture}
\begin{axis}[
  width=0.47\linewidth, height=5.2cm, ymode=log,
  xlabel={$H_{\mathrm{uni}}$ (nats)}, ylabel={gen-PPL},
  xmin=1.9, xmax=4.8, ymin=0.9, ymax=900,
  tick label style={font=\scriptsize}, label style={font=\scriptsize},
  title style={font=\scriptsize}, title={(a) LM1B, $L{=}128$},
  grid=major, grid style={gray!20}]
\fill[blue!8] (axis cs:4.13,0.9) rectangle (axis cs:4.39,900);
\addplot[black,line width=1.0pt,forget plot] coordinates {(1.9,56.9)(4.8,56.9)};
\node[font=\tiny,anchor=north west] at (axis cs:1.95,52) {reference gen-PPL};
\addplot[red!70!black,mark=*,mark size=1.1pt] coordinates
  {(2.54,39.79)(2.99,75.00)(3.33,117.37)(3.59,181.88)(3.78,286.01)(3.96,460.03)(4.11,762.63)};
\addplot[orange!85!black,mark=square*,mark size=1.1pt] coordinates
  {(2.12,9.10)(3.02,14.98)(3.24,18.11)(3.56,29.00)(3.67,36.89)(3.84,61.26)};
\addplot[violet,mark=triangle*,mark size=1.4pt] coordinates
  {(2.08,1.70)(2.48,2.02)(2.77,2.47)(2.99,3.14)(3.47,5.63)(3.89,13.86)(4.16,29.39)(4.55,234.13)};
\addplot[teal!80!black,mark=diamond*,mark size=1.4pt] coordinates
  {(3.10,10.33)(3.32,27.78)(3.51,39.59)(3.68,49.33)(3.82,61.82)(3.96,65.20)(4.03,78.13)(4.10,88.68)(4.16,107.86)};
\end{axis}
\end{tikzpicture}\hfill
\begin{tikzpicture}
\begin{axis}[
  width=0.47\linewidth, height=5.2cm, ymode=log,
  xlabel={$H_{\mathrm{uni}}$ (nats)}, ylabel={gen-PPL},
  xmin=2.1, xmax=6.2, ymin=1.5, ymax=700,
  tick label style={font=\scriptsize}, label style={font=\scriptsize},
  title style={font=\scriptsize}, title={(b) OpenWebText, $L{=}1024$},
  grid=major, grid style={gray!20}]
\fill[blue!8] (axis cs:5.18,1.5) rectangle (axis cs:5.64,700);
\node[font=\tiny,blue!45!black,anchor=south] at (axis cs:5.41,330)
  {trained range};
\addplot[black,line width=1.0pt] coordinates {(2.1,17.1)(6.2,17.1)};
\node[font=\tiny,anchor=north west] at (axis cs:2.15,15.6) {reference gen-PPL};
\addplot[red!70!black,mark=*,mark size=1.1pt] coordinates
  {(3.33,59.46)(3.78,99.88)(4.17,152.88)(4.51,231.23)(4.81,346.93)(5.09,543.63)};
\addplot[orange!85!black,mark=square*,mark size=1.1pt] coordinates
  {(2.21,19.32)(3.36,19.35)(3.77,22.08)(4.48,29.49)(4.73,34.35)(5.14,50.55)};
\addplot[violet,mark=triangle*,mark size=1.4pt] coordinates
  {(2.77,3.57)(3.47,2.25)(4.16,2.10)(4.60,2.52)(5.01,3.36)(5.30,4.73)(5.41,5.79)(5.55,7.14)(5.69,9.90)(5.97,21.63)};
\addplot[teal!80!black,mark=diamond*,mark size=1.4pt] coordinates
  {(2.54,3.89)(2.72,5.54)(3.55,12.03)(3.80,16.59)(3.89,22.63)(4.20,32.71)(4.48,49.64)(4.63,58.84)(4.63,60.74)};
\end{axis}
\end{tikzpicture}
\\[2pt]
{\scriptsize
\textcolor{red!70!black}{$\bullet$}~Top-$k$ IID \quad
\textcolor{orange!85!black}{$\blacksquare$}~Mirror-$k$ \quad
\textcolor{violet}{$\blacktriangle$}~Periodic-$k$ \quad
\textcolor{teal!80!black}{$\blacklozenge$}~Phrase bank-$m$}
\caption{Measured parameter sweeps for the four training-free samplers in
the gen-PPL/entropy plane. On OpenWebText, Periodic configurations within
the trained generators' entropy range (shaded) have measured gen-PPL
$4.73$--$7.14$, compared with $17.1$ for reference text (solid line).
Table~\ref{tab:rebench} reports OpenWebText Periodic-$400$ at gen-PPL $19.6$
and unigram entropy $5.93$. Every plotted point is a measured bank.}
\label{fig:frontier}
\end{figure}

% =========================================================================
\section{Jeffreys coding divergence}
\label{app:ctw}

In this appendix, we detail the construction and estimation of Jeffreys coding
divergence (JCD). We derive the estimator from the coding interpretation of the
Kullback--Leibler divergence, describe its instantiation via frozen
context-tree weighting, and analyze directional coverage and fidelity results
alongside the estimator's structural limitations.

\subsection{Estimator}

In information theory, the Kullback--Leibler divergence has an exact operational
meaning: it measures the expected coding overhead incurred when data generated by
one distribution are compressed using an optimal code designed for another. If
both the data distribution \(P(x)\) and the generator's distribution \(G(x)\)
could be evaluated directly, their directional cross-entropies would immediately
isolate coverage and fidelity mismatches:
\begin{align}
H_\times(P,G) - H(P) &= D_{\mathrm{KL}}(P\|G), &&\text{coverage mismatch},\label{eq:cov}\\
H_\times(G,P) - H(G) &= D_{\mathrm{KL}}(G\|P), &&\text{fidelity mismatch}.\label{eq:fid}
\end{align}
Coverage mismatch measures the penalty for mode dropping (how poorly the
generator encodes real human text), whereas fidelity mismatch measures the
penalty for hallucinating improbable text (how poorly the data distribution
encodes generated samples). The symmetric Jeffreys divergence is their arithmetic
mean.

Because neither \(P(x)\) nor the diffusion likelihood \(G(x)\) is tractable in
practice, we replace them with empirical distribution estimates. We fit an
external universal compression algorithm \(C\) separately on training splits
from each source and freeze the resulting coders:
\begin{equation}
P_{\mathrm{fit}} \xrightarrow{\;C\;} q_P,
\qquad
G_{\mathrm{fit}} \xrightarrow{\;C\;} q_G.
\label{eq:fitfreeze}
\end{equation}
We then evaluate held-out sample banks \(\mathcal{B}_P\) and \(\mathcal{B}_G\)
under both frozen coders to estimate directional cross-coding overhead:
\begin{equation}
\widehat{D}_{\mathrm{cov}}
= \frac{\sum_{x \in \mathcal{B}_P} \log_2 \frac{q_P(x)}{q_G(x)}}
       {\sum_{x \in \mathcal{B}_P} |x|_{\mathrm{bytes}}},
\qquad
\widehat{D}_{\mathrm{fid}}
= \frac{\sum_{x \in \mathcal{B}_G} \log_2 \frac{q_G(x)}{q_P(x)}}
       {\sum_{x \in \mathcal{B}_G} |x|_{\mathrm{bytes}}},
\qquad
\mathrm{JCD} = \tfrac{1}{2} \bigl( \widehat{D}_{\mathrm{cov}} + \widehat{D}_{\mathrm{fid}} \bigr).
\label{eq:cov-hat}
\end{equation}
Evaluating each sample under both coders provides a crucial normalization:
because \(q_P\) and \(q_G\) share identical inductive biases and model capacity,
document-specific compressibility and universal coding overhead largely cancel
out in the log-ratio. The resulting estimates are byte-normalized, finite-context
surrogates for Equations~\eqref{eq:cov}--\eqref{eq:fid}, pooled across documents
so that longer texts contribute proportionally to their length. We write
\(\mathrm{JCD}_{\rightarrow} = \widehat{D}_{\mathrm{cov}}\) and
\(\mathrm{JCD}_{\leftarrow} = \widehat{D}_{\mathrm{fid}}\);
Table~\ref{tab:rebench} reports their symmetric mean.

\subsection{Frozen context-tree weighting}

\begin{figure}[!htbp]
\centering
\includegraphics[width=0.92\linewidth]{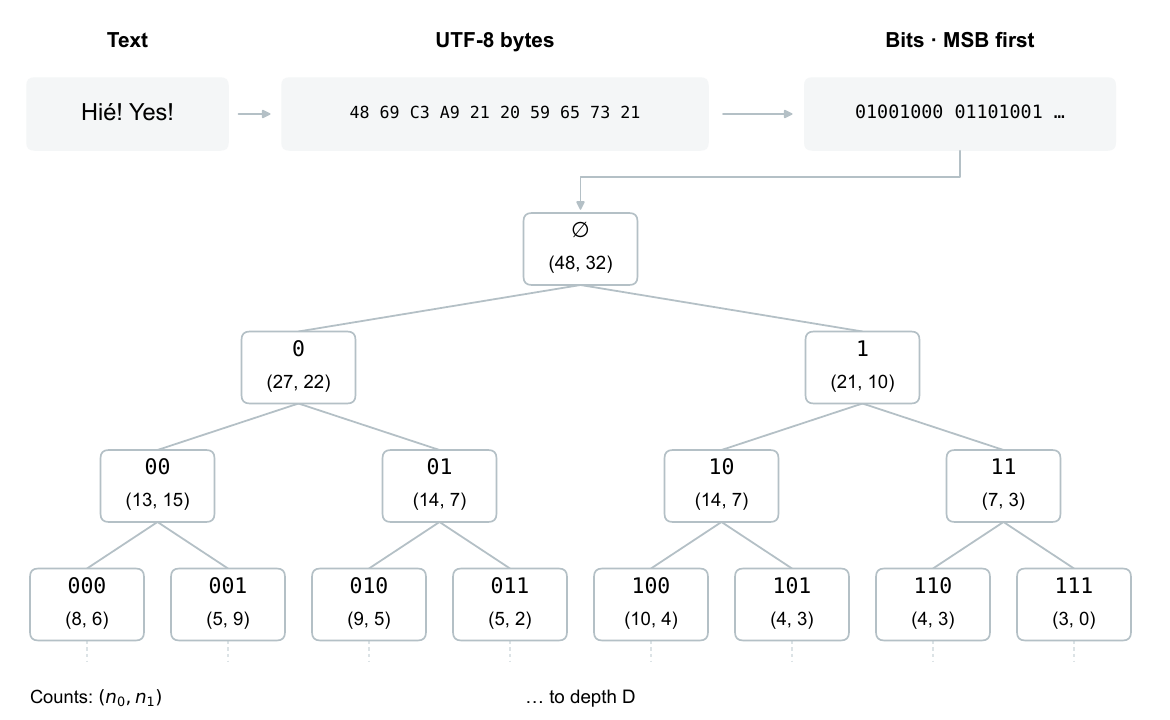}
\caption{\textbf{Fitting.} One source is read as a raw UTF-8 bit stream and
used to grow a context tree. Each node is labelled by the bit context that
reaches it---$\varnothing$ at the root, then $0$ and $1$, then $00,01,10,11$,
and so on---and stores the counts $(n_0,n_1)$ of the bits observed in that
context, from which it forms a Krichevsky--Trofimov estimate. }
\label{fig:ctw-fit}
\end{figure}

We instantiate the universal learner \(C\) using context-tree weighting
(CTW; \citealp{willems1995context}) over raw UTF-8 byte streams expanded into
bits, most significant bit first (Figure~\ref{fig:ctw-fit}). Operating at the bit
level avoids any reliance on tokenizers or vocabulary assumptions.

A classic challenge in statistical language modeling is choosing the Markov
order: short contexts miss local syntax, while deep contexts suffer from severe
sample sparsity. CTW resolves this dilemma elegantly. Rather than committing to
a single heuristic context length, it constructs a complete suffix tree up to a
maximum depth \(D\) and computes an exact Bayesian mixture over all \(2^{2^D}\)
possible tree-pruning structures in \(O(D)\) time per bit.

We configure CTW at depth \(D=20\) bits (corresponding to at most \(2.5\) bytes
of past context). Each context string \(s\) of length \(|s| \le D\) maintains
observed bit counts \((n^s_0, n^s_1)\), with context histories reset at document
boundaries. At each node, conditional bit probabilities are estimated via the
Krichevsky--Trofimov (KT) estimator~\citep{krichevsky1981performance}, which uses
a minimax-optimal \(\operatorname{Beta}(\tfrac{1}{2}, \tfrac{1}{2})\) prior:
\begin{equation}
P^{s}_{\mathrm{kt}}(b) = \frac{n^s_b + \tfrac{1}{2}}{n^s_0 + n^s_1 + 1},
\qquad
\bar{P}^{s}_{\mathrm{kt}}
= \frac{\Gamma(\tfrac{1}{2} + n^s_0)\,\Gamma(\tfrac{1}{2} + n^s_1)}
       {\Gamma(\tfrac{1}{2})^2\,\Gamma(1 + n^s_0 + n^s_1)}.
\label{eq:kt}
\end{equation}
The weighted block likelihood \(P^s_{\mathrm{w}}\) combines the node's local
predictive model with the weighted predictions of its child subtrees:
\(P^{s}_{\mathrm{w}} = \tfrac{1}{2}\bar{P}^{s}_{\mathrm{kt}} + \tfrac{1}{2}P^{s0}_{\mathrm{w}}P^{s1}_{\mathrm{w}}\)
(with \(P^{s}_{\mathrm{w}} = \bar{P}^{s}_{\mathrm{kt}}\) at the leaf depth \(D\)).
The posterior stopping weight,
\(\beta_s = \tfrac{1}{2}\bar{P}^{s}_{\mathrm{kt}} / P^{s}_{\mathrm{w}}\),
measures the posterior probability that the true context tree stops at node \(s\).
We compute \(\beta_s\) once during the fitting phase and then freeze it.

To evaluate held-out text without adapting the model during scoring, we combine
predictions along the active context path from the leaf to the root:
\begin{equation}
q^{(D)}_t(b) = P^{s^{(D)}_t}_{\mathrm{kt}}(b),
\qquad
q^{(d)}_t(b) = \beta_{s^{(d)}_t}P^{s^{(d)}_t}_{\mathrm{kt}}(b)
+ \bigl(1 - \beta_{s^{(d)}_t}\bigr)q^{(d+1)}_t(b).
\label{eq:ctw-predictive}
\end{equation}
The predictive probability \(q(b_t \mid b_{<t}) = q^{(0)}_t(b_t)\) is thus a
smooth mixture across all \(D+1\) context depths. The total code length
\(-\log_2 q(x) = -\sum_t \log_2 q(b_t \mid b_{<t})\) is accumulated bit by bit
with \(O(D)\) computational cost per bit.

Freezing the context trees is a vital methodological choice. Standard online CTW
updates node counts adaptively as bits are processed. In an evaluation setting,
however, online adaptation would allow each coder to learn the target evaluation
bank on the fly, artificially shrinking the cross-entropy gap. Freezing ensures
that the coder acts purely as a static reference density.
To illustrate the recursion concretely, consider a toy tree at depth \(D=2\)
with counts \(\varnothing{:}(5,3)\), \(0{:}(3,1)\), and \(00{:}(2,1)\). The
first bit of a candidate string \(x=010\) receives predictive probability
\(q(0) = 0.6296\), formed as a weighted mixture of node predictives
\((0.611, 0.700, 0.625)\) with weights \((0.667, 0.185, 0.148)\). Our
depth-20 implementation executes this identical Bayesian weighting across 21
levels.

\subsection{Directional results}

\begin{figure}[!htbp]
\centering
\includegraphics[trim=42bp 3bp 46bp 13bp,clip,width=0.64\linewidth,height=0.43\textheight,keepaspectratio]{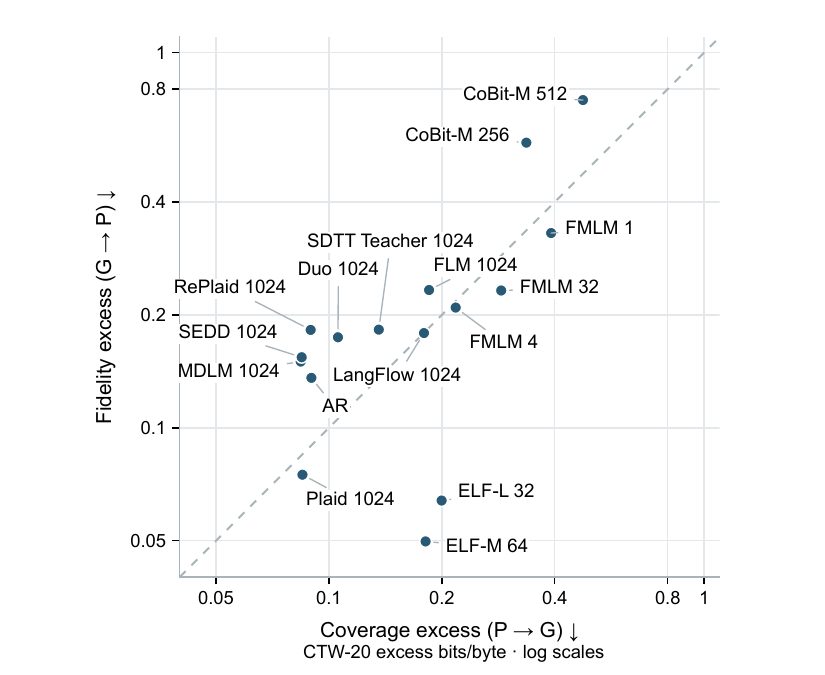}\\[3pt]
\includegraphics[trim=42bp 3bp 46bp 13bp,clip,width=0.64\linewidth,height=0.43\textheight,keepaspectratio]{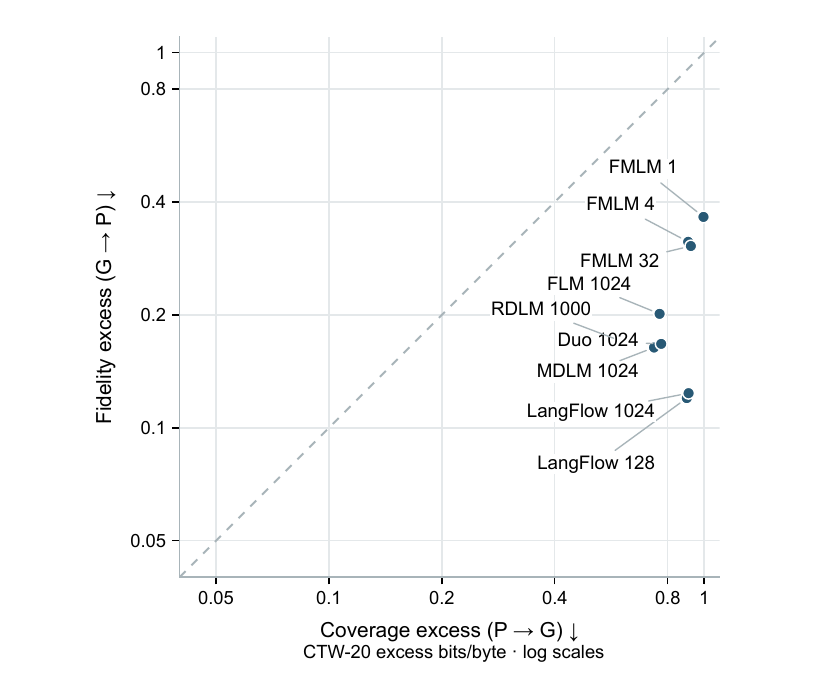}
\caption{Directional Jeffreys coding divergence for trained configurations on
OpenWebText ($L{=}1024$, top) and LM1B ($L{=}128$, bottom), with identical
log axes. Horizontal: local coverage mismatch; vertical: local fidelity
mismatch. Numbers in labels denote NFE.}
\label{fig:coverage-fidelity-trained}
\end{figure}

A key strength of JCD is that resolving the symmetric divergence into its
constituent directions reveals structural asymmetries that a single scalar
obscures. Figure~\ref{fig:coverage-fidelity-trained} plots directional JCD
across trained generators on OpenWebText and LM1B.

Points along the diagonal indicate balanced models, such as Plaid, which achieves
nearly identical coverage (\(0.085\) bits/byte) and fidelity (\(0.075\)
bits/byte) overhead. In contrast, models situated off the diagonal expose
specific modeling trade-offs. For example, ELF-M at 64 NFE exhibits very low
fidelity excess (\(0.050\) bits/byte) alongside substantial coverage excess
(\(0.181\) bits/byte). This asymmetry indicates that while its generated samples
appear highly realistic to the human-text coder, its model fails to capture the full
variety of real text—a characteristic signature of mode dropping.

Crucially, both directional components decisively reject our training-free
heuristics. The failure is especially stark along the coverage axis: OpenWebText
Periodic-400 and Top-\(k\) IID incur massive coverage penalties of \(5.94\) and
\(7.85\) bits/byte, respectively, compared to at most \(0.48\) bits/byte for any
trained generator. JCD thus exposes degenerate text immediately without requiring
neural representations or auxiliary classifiers.

\subsection{Limitations}

While JCD provides an encoder-free evaluation signal, its design involves two
clear trade-offs:
\begin{itemize}[leftmargin=1.4em,itemsep=2pt,parsep=0pt,topsep=2pt]
\item \textbf{Finite context horizon:} A context tree of depth \(D=20\) bits
conditions each prediction on at most \(2.5\) bytes of history. It serves as a
rigorous test of local morphology, orthography, and n-gram syntax, but is
naturally blind to long-range semantic coherence, narrative progression, and
discourse structure. A favorable JCD score confirms local distributional
agreement, not global textual coherence.
\item \textbf{High-order context sparsity:} At depth \(D=20\), the context tree
contains up to \(2^{20} \approx 10^6\) possible leaf contexts. On finite training
splits of 512 documents, many deep contexts remain unobserved, causing node
predictions to revert to the uninformative \(\tfrac{1}{2}\) Beta prior. This
sparsity affects real and generated coders differently depending on vocabulary
concentration. Increasing \(D\) exacerbates this sparsity without adding
meaningful statistical signal.
\end{itemize}
Consequently, JCD measurements should be interpreted as finite-context empirical
surrogates rather than certified mathematical bounds on true full-sequence KL
divergences. Absolute bits-per-byte differences are comparable only under identical
coder configurations, and JCD is most informative when paired with high-level
semantic metrics such as MAUVE and Energy Distance.

% =========================================================================
\section{Metric sensitivity}
\label{app:sensitivity}

Every evaluation metric embodies implicit assumptions regarding sample budgets,
evaluation cohorts, and feature representations. In this appendix, we
systematically stress-test these dependencies: analyzing how sample count
governs MAUVE, how metric correlations shift when adversarial heuristics are
excluded, and whether model rankings are driven primarily by the chosen
mathematical distance or the underlying neural representation.

\subsection{MAUVE and the number of scored samples}

Distributional estimators over continuous feature spaces are inherently sensitive
to finite-sample effects. For MAUVE~\citep{mauve2021}, which quantizes learned
features into \(k\)-means clusters to approximate divergence curves, sample size
\(n\) directly determines the resolution of the empirical partition. While
Table~\ref{tab:rebench} standardizes on \(n=1{,}024\) samples per side, published
benchmarks often report MAUVE under much smaller sample budgets.

\begin{figure}[!htbp]
\centering
\includegraphics[width=0.84\linewidth]{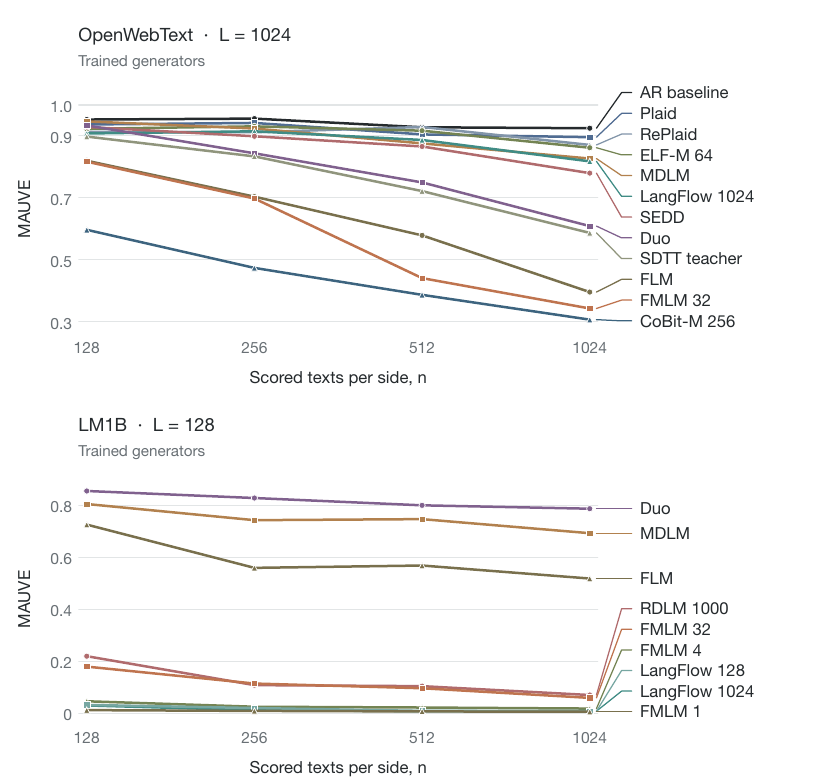}
\caption{MAUVE sensitivity to scored sample count for trained generators on
OpenWebText (top) and LM1B (bottom) from Table~\ref{tab:rebench}. Points
average five fixed subset seeds with $n$ reference and $n$ generated texts.
Panels use their labeled y-axis ranges.}
\label{fig:mauve-scored-count}
\end{figure}

Figure~\ref{fig:mauve-scored-count} illustrates the consequences of varying
\(n \in \{128, 256, 512, 1{,}024\}\). Two systematic patterns emerge:
\begin{itemize}[leftmargin=1.4em,itemsep=2pt,parsep=0pt,topsep=2pt]
\item \textbf{Optimism under small budgets:} Smaller sample counts produce
consistently inflated scores. Every evaluated generator scores higher at
\(n=128\) than at \(n=1{,}024\), while the score spread contracts substantially
(from \(0.62\) down to \(0.36\) on OpenWebText). Coarse partitions overestimate
distributional overlap.
\item \textbf{Rank stability depends on corpus diversity:} While generator
rankings remain remarkably stable on LM1B (pairwise Spearman \(\rho \ge 0.98\)),
smaller sample sizes induce noticeable rank churn on OpenWebText (\(\rho = 0.81\)
between \(n=128\) and \(1{,}024\)). Closely scored models frequently swap positions;
for instance, Duo drops from 4th to 8th as \(n\) increases.
\end{itemize}
These scaling dynamics confirm that MAUVE scores cannot be directly compared
across studies unless evaluated under identical sample counts.

\subsection{Metric agreement across generators}
\label{app:trained-agreement}

\begin{figure}[!htbp]
\centering
\includegraphics[trim=31bp 18bp 14bp 23bp,clip,width=0.82\linewidth]{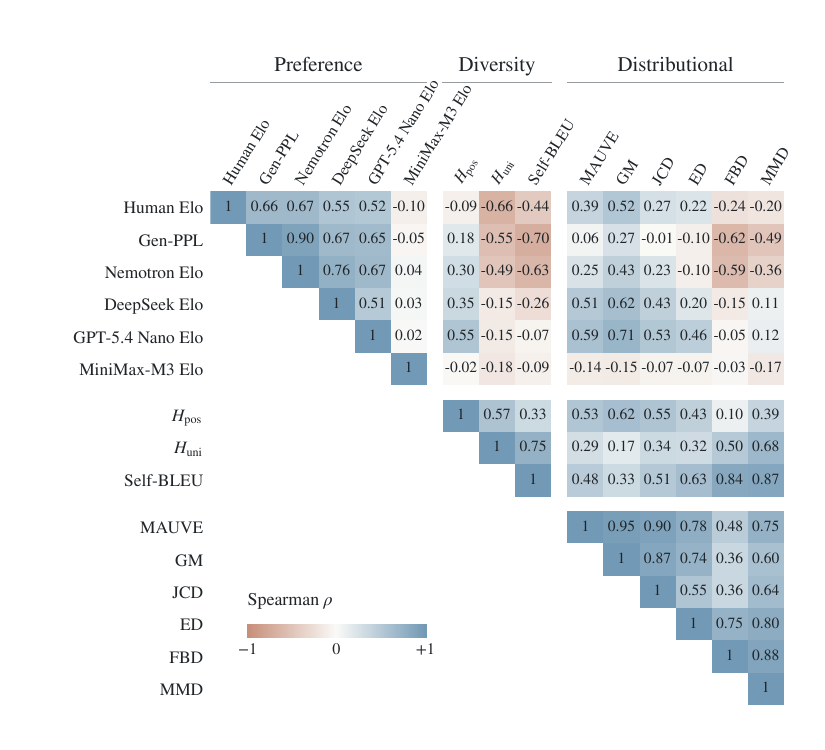}
\caption{Metric agreement over the 12 trained OpenWebText generators, excluding
the naive samplers and the real-text reference. Direction-corrected tied-rank
Spearman \(\rho\), computed as in Figure~\ref{fig:agreement-all}.
Human Elo uses the final 300-comparison snapshot.}
\label{fig:agreement-trained}
\end{figure}

Our primary evaluation matrix in Section~\ref{sec:rebench}
(Figure~\ref{fig:agreement-all}) joins the 12 trained generators with the four
training-free heuristics. Because the heuristics were constructed specifically to
exploit \(\genppl\), their extreme scores inevitably exert strong leverage on
overall rank correlations. To evaluate how metrics align under standard
benchmarking conditions, Figure~\ref{fig:agreement-trained} recomputes the
correlation matrix exclusively over the 12 trained OpenWebText generators.

Restricting the cohort to trained models reveals a striking divergence:
\begin{itemize}[leftmargin=1.4em,itemsep=2pt,parsep=0pt,topsep=2pt]
\item \textbf{Likelihood tracks fluency in non-adversarial text:} Within trained
models, \(\genppl\) correlates strongly with LLM preference judgments (Spearman
\(\rho = 0.90\) with Nemotron Elo, up from \(0.46\) across the full cohort). In
the absence of degenerate repetition, higher scorer likelihood generally aligns
with grammatical surface fluency.
\item \textbf{Likelihood diverges from distributional fidelity:} Despite its
alignment with preference judges, \(\genppl\) correlates negatively with
distributional discrepancies (FBD and MMD) as well as unigram entropy and
Self-BLEU.
\item \textbf{Distributional agreement remains robust:} The distributional
metrics (MAUVE, Gradient Moment, Energy Distance, and FBD) maintain robust
mutual agreement (\(\rho = 0.36\)--\(0.95\)).
\end{itemize}
This contrast reinforces our central finding: even when generators do not
actively exploit the scorer, optimizing for sample likelihood under a frozen
model remains fundamentally distinct from matching the real data distribution.

\subsection{Encoder-space sensitivity}
\label{app:encoder-sensitivity}

A foundational question for feature-based evaluation is whether model rankings
reflect the chosen statistical distance or the geometric properties of the
underlying embedding space. To isolate these factors, we evaluate four distinct
distance metrics—MAUVE, Gaussian Fr\'echet distance, Euclidean energy distance,
and kernel \(\mathrm{MMD}^2\)—across five frozen encoder architectures:
GPT-2-large, BERT-base-uncased~\citep{devlin2019bert},
RoBERTa-base~\citep{liu2019roberta}, all-MPNet-base-v2~\citep{allmpnetbasev2}, and
E5-base-v2~\citep{wang2022e5}.

Each pipeline applies its native tokenizer and pooling rule (capped at 512
tokens), L2-normalizes document vectors, and fits PCA whitening strictly on
reference text to project representations down to 128 dimensions.

\begin{figure}[!htbp]
\centering
\includegraphics[width=0.98\linewidth]{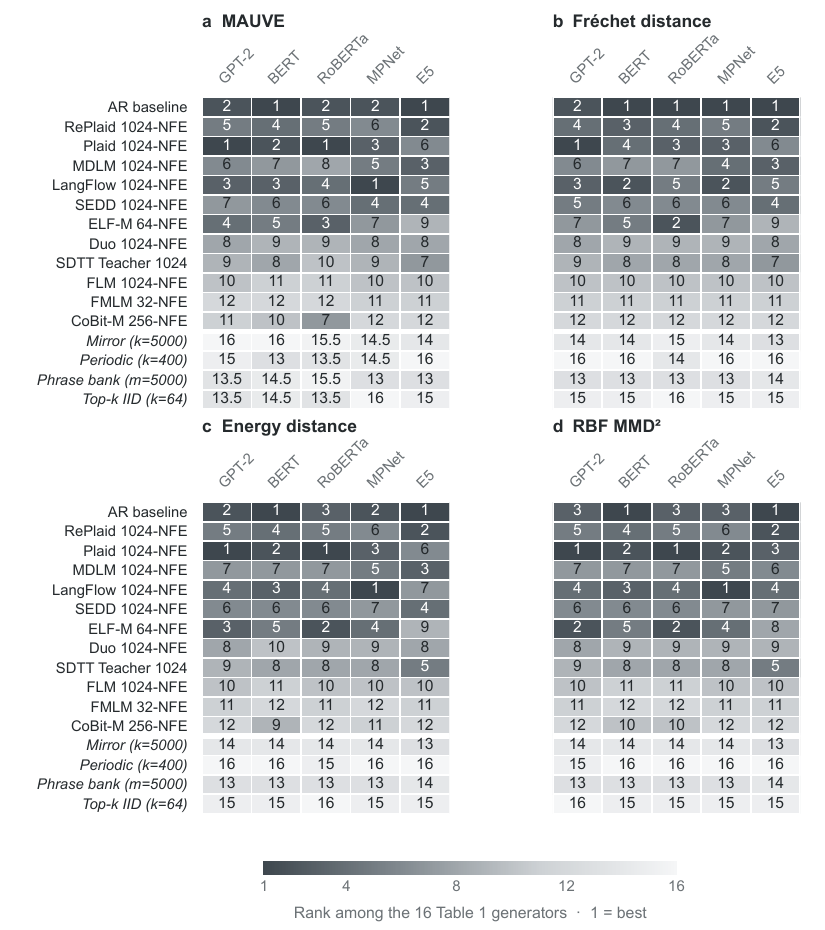}
\caption{Encoder-space sensitivity of four feature-space metrics on OpenWebText:
(a) MAUVE, (b) squared Gaussian Fr\'echet distance, (c) unbiased Euclidean
energy distance, and (d) unbiased three-bandwidth Gaussian RBF MMD$^2$.
Rows show the 16 OpenWebText generators in Table~\ref{tab:rebench}.
Each cell gives a rank within this cohort (1 best), with average ranks for
exact ties; darker shading indicates a better rank. Rankings use
1,024 reference and 1,024 generated texts per score. MAUVE is ranked high to
low; the other three scores are ranked low to high.}
\label{fig:encoder-sensitivity}
\end{figure}

\begin{figure}[!htbp]
\centering
\includegraphics[width=\linewidth]{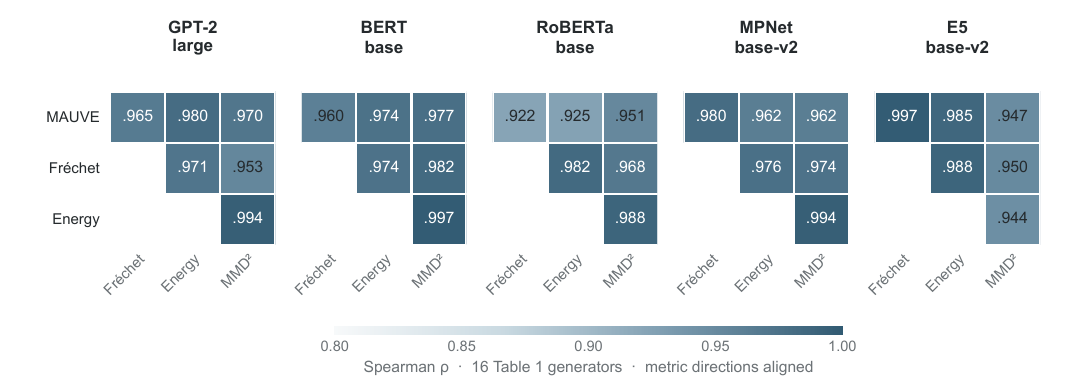}
\caption{Within-encoder metric agreement on OpenWebText. Each upper triangle
gives pairwise Spearman correlations across the 16 generators in
Table~\ref{tab:rebench} (12 trained and four constructed), excluding the
real-text reference. Metric directions are aligned and ties receive average
ranks; self-correlations and duplicate pairs are omitted.}
\label{fig:encoder-metric-correlations}
\end{figure}

Our experiments reveal a clear hierarchy: \emph{the representation matters far
more than the distance function}. As documented in
Figure~\ref{fig:encoder-metric-correlations}, for any fixed encoder, the four
mathematical distances produce nearly identical generator rankings (pairwise
Spearman \(\rho = 0.922\)--\(0.997\), mean \(0.970\)). In contrast,
Figure~\ref{fig:encoder-sensitivity} shows that swapping the encoder pipeline
induces substantial rank shifts across several trained generators. Because
different encoders retain distinct linguistic, syntactic, and semantic
invariances, the choice of representation is the dominant factor determining
which distributional discrepancies a metric can detect.

% =========================================================================
\section{Pairwise judgments}
\label{app:judges}

While automated distributional metrics capture statistical divergence, they
cannot directly assess subjective readability or perceptual quality. In this
appendix, we detail our two pairwise evaluation campaigns: an automated study
using four distinct large language model judges, and a human preference study
conducted on excerpted prefixes.

\subsection{LLM judges}
\label{app:llm-judges}
\label{app:other-judges}% retained for existing cross-references

Pairwise evaluation via LLM-as-a-judge has become a standard approach for
assessing generated text, but reported outcomes depend sensitively on the judge
model, prompt design, and decision protocol~\citep{zheng2023llmjudge}. Rather
than relying on a single model, we evaluate four distinct LLM judges under an
identical blind protocol: Nemotron-3-Ultra-550B-A55B~\citep{nvidia2026nemotronultra},
DeepSeek, GPT-5.4 Nano~\citep{openai2026gpt54nano}, and
MiniMax-M3~\citep{minimax2026m3}.

Each judge receives two anonymous, full-length generations and is prompted to
evaluate readability, coherence, naturalness, and overall writing quality.
Sample text is treated strictly as untrusted input, and the judge outputs a
single JSON decision among four options: \texttt{a\_better}, \texttt{b\_better},
\texttt{both\_good}, and \texttt{both\_bad}.

To control for position bias (the tendency to favor whichever text is presented
first), every evaluated pair is scored in both presentation orders across a
shared schedule of \(16{,}560\) matches. This campaign covers all 21 frozen
OpenWebText banks, including the real-text reference. After canonicalizing
order, a decisive win scores \(s=1\) (or \(0\)), while either tie option scores
\(s=\tfrac{1}{2}\). Ratings initialize at 1500 and update sequentially in a
fixed, content-deterministic schedule via standard Elo:
\begin{equation}
E_{ij} = \bigl(1 + 10^{(R_j - R_i)/400}\bigr)^{-1},
\qquad
R_i \leftarrow R_i + 32\,(s - E_{ij}),
\qquad
R_j \leftarrow R_j - 32\,(s - E_{ij}).
\label{eq:elo-update}
\end{equation}
Because Elo is a zero-sum, order-dependent rating algorithm, scores reflect
relative win rates within this closed cohort rather than absolute capability,
and numerical scales cannot be compared directly across different judges.
Table~\ref{tab:judges} displays the resulting ratings for the 17 OpenWebText
banks reported in Table~\ref{tab:rebench}, retained from the complete 21-bank
campaign.

The four judges differ most dramatically in their willingness to declare a
winner:
\begin{itemize}[leftmargin=1.4em,itemsep=2pt,parsep=0pt,topsep=2pt]
\item \textbf{High hesitancy compresses ratings:} MiniMax-M3 declines to choose
a winner in 93\% of comparisons, defaulting almost entirely to \texttt{both\_bad}.
DeepSeek and GPT-5.4 Nano record tie rates of 58\%. Because every tie yields
\(s=\tfrac{1}{2}\), high hesitancy prevents ratings from separating:
MiniMax-M3's scores span a narrow band of 1469--1532, amplifying noise and
producing erratic rankings (such as ranking Top-\(k\) IID third while placing
the autoregressive baseline twentieth).
\item \textbf{Decisive judging separates models:} In contrast, Nemotron acts as
a decisive evaluator, recording a tie rate of only 22\% and producing a wide,
informative rating spread across generators.
\end{itemize}
These behavioral differences demonstrate why preference scores from distinct
models should not be averaged into a composite score: doing so conflates
decisive judgments with uncalibrated or overly conservative evaluators. We
therefore report each judge separately and interpret MiniMax-M3 with caution.

For the 16 generated OpenWebText banks in Table~\ref{tab:rebench}, we join
ratings to metric values using exact matches on dataset, generator name, and
bank digest. Spearman rank correlations treat higher ratings as better and
assign average ranks to ties. The full prompt template and per-step rating
traces are released alongside our evaluation code.

\subsection{Human comparisons}
\label{app:human-judges}

To calibrate whether automated LLM judges reflect human perception, we
conducted a targeted human evaluation across the twelve trained OpenWebText
generators and the real-text reference. We deliberately excluded the
training-free heuristics from this study, as their incoherence is obvious upon
inspection; the human study is designed specifically to resolve distinctions
among trained models.

Raters were shown anonymous 35-word prefixes from two competing sources and
instructed to select the excerpt with superior fluency, naturalness, and
coherence (or skip if undecided). Skips leave ratings unaffected, while
decisions update Elo scores sequentially according to
Equation~\eqref{eq:elo-update}.

All reported human ratings use a frozen snapshot collected on 25 September 2026.
Nine independent raters completed 304 comparisons; removing four flagged
entries leaves 300 verified judgments. These cover all 78 possible source pairs,
averaging 41--49 matches per generator. The resulting leaderboard orders the
generators as follows: ELF-M, Plaid, Reference, LangFlow, SDTT Teacher, AR
baseline, FLM, RePlaid, FMLM, CoBit-M, Duo, MDLM, and SEDD
(Table~\ref{tab:human-elo}).

This study provides an essential coarse calibration, but its design involves
two necessary trade-offs:
\begin{itemize}[leftmargin=1.4em,itemsep=2pt,parsep=0pt,topsep=2pt]
\item \textbf{Local prefixes versus full documents:} Humans evaluated 35-word
prefixes to maintain rater throughput and attention across 78 pairings. This
protocol reliably captures immediate surface fluency, grammatical consistency,
and local naturalness, but is naturally blind to long-range narrative
coherence and global thematic progression.
\item \textbf{Finite sample scale:} With 41--49 comparisons per generator,
individual ratings carry noticeable variance. Adjacent models differ by only a
few Elo points and should be viewed as statistically comparable rather than
strictly ordered.
\end{itemize}
Furthermore, because human raters evaluated short excerpts while LLM judges
assessed complete 1,024-token texts, absolute Elo values cannot be compared
between the two campaigns. Human Elo serves as a valuable directional sanity
check for our automated profile, confirming that while no single metric is
flawless, our multi-faceted profile broadly reflects human perception.

% =========================================================================
\FloatBarrier
\section{Text Samples}
\label{app:gallery}

To provide qualitative context for our quantitative benchmarks, this appendix
presents raw generated samples across all evaluated model families, reference
corpora, and training-free heuristics.

To prevent cherry-picking, every displayed sample is selected uniformly at
random from its frozen \(1{,}024\)-sample benchmark bank without manual
inspection or post-filtering. Samples are rendered with tokenizer artifacts
intact---including special delimiters such as \gallerytoken{<|endoftext|>} or
\gallerytoken{[CLS]} and replacement characters \gallerytoken{[U+FFFD]}---to
transparently reflect each model's raw decoding output.

These qualitative samples illustrate the core paradox examined in this paper:
trained diffusion language models produce fluent, grammatically structured
English, whereas heuristic samplers produce ungrammatical word lists or periodic
loops. Yet under standard \(\genppl\) and unigram entropy, these heuristics
match or outperform the trained models. Comparing these cards visually makes
the failure of single-metric likelihood evaluation immediately apparent.

\subsection{OpenWebText samples}
\label{app:gallery-owt}

The following cards display samples generated on OpenWebText at context length
\(L=1{,}024\) tokens, truncated to their leading excerpts to fit a compact paired
layout. Models are presented in the order of Table~\ref{tab:rebench}, alongside
reference human text and the four training-free heuristics.

\gallerysingle{%
% Original bank metadata (not printed): \textbf{Reference}\newline \nolinkurl{owt_data_train}\newline \emph{sample owt-owt\_data\_train-000477}\newline Gen-PPL$\downarrow$: 17.1; $H_{\rm emp}\uparrow$: 5.5
\gallerycard{Reference}{}{477}{%
\gallerytoken{<|endoftext|>} before I headed out I put the girls in their Vikings jammies and got ready to leave. Before I left I asked my older daughter a very important question: "Are the Vikings gonna win tonight Gemma?" "Yep! Skoohhh Vikings, less Gooo!" And then she blew me a kiss goodbye. Just like the Packers kissed the NFC North title goodbye.\gallerytoken{<|endoftext|>}Sprites: A Rarely Seen Sky Phenomenon Caught On Camera When thunderstorms emit lightning, we see the white, snaking electricity from the ground. But if you flew above the clouds, you would see a sky phenomenon known as sprites. These are rarely seen bolts of red light that look like very fast burning sparklers. The Capital Weather Gang over at The Washington Post describes them like this: "[Sprites] can reach 50-60 miles into space and penetrate downward into the middle of the stratosphere (15-20 miles high) with jellyfish-like tendrils." [...]
}
}
\gallerypair{%
% Original bank metadata (not printed): \textbf{AR baseline \hfill NFE=1}\newline \nolinkurl{owt_ar_base}\newline \emph{sample 578}\newline Gen-PPL$\downarrow$: \missingresult; $H_{\rm emp}\uparrow$: \missingresult
\gallerycard{AR baseline}{NFE\,=\,1\,024}{578}{%
\gallerytoken{<|endoftext|>} have clean, inexpensive machines packed full of the most powerful computer devices you can imagine --- all operated in a simple and efficient manner that involves little risk. Plus, why not choose a PC that has the ease of integration with existing plumbing (or mesh, as some will call it) and is equipped with 4K video cameras? Ty to the television, then. Generally speaking, the purest multimedia vinyl The One (I guess because that's what it is) died before it got to the last page. And since Microsoft didn't deliver, those rolls of vinyl---which I happen to like, at least verging on sentimental---aren't going to be available to purchase on October 25th anyway. But hey, at least you'll be able to get out there and sample some tunes for two years. Oh, and it's less costly to upgrade to a Surface Pro 3 with Kinect gadgets. BeCaClkOojckPPJmbMPur. How expensive is the Surface [...]
}
}{%
% Original bank metadata (not printed): \textbf{RePlaid (no-SC, DDPM) \hfill NFE=1024}\newline \nolinkurl{owt_v2_replaid_nosc_ddpm_1024_nfe}\newline \emph{sample 123}\newline Gen-PPL$\downarrow$: \missingresult; $H_{\rm emp}\uparrow$: \missingresult
\gallerycard{RePlaid (no-SC, DDPM)}{NFE\,=\,1\,024}{123}{%
\gallerytoken{<|endoftext|>} Indonesia, rejected an unprecedented action by the emirate to imbue Burundi with the Right of Return in response to the public and the controversial question of whether the historical event should be celebrated, wrote Arthur Duncan Snoville. Rapovano resigned from the Honya and government of Burundi in February 2014, stating that he wants Archbishop Dutton to give him "pay" for his actions in the past. The widely published islands list includes hurricanes, but the Honya's effective lobbying is making their case to a global public including Pope Benedict XVI. Archbishop Dolan's comments were aimed at connecting Indonesia's largest island nation with the French-controlled territory to the east of the Sahara, in a similar fashion through parts of the Pacific. The Honya's message is broader, however, which reinforces the surroundings contributed to and considered the source of [...]
}
}

\gallerypair{%
% Original bank metadata (not printed): \textbf{Plaid \hfill NFE=1024}\newline \nolinkurl{owt_v2_plaid_1024_nfe}\newline \emph{sample owt-owt\_v2\_plaid\_1024\_nfe-000445}\newline Gen-PPL$\downarrow$: 20.7; $H_{\rm emp}\uparrow$: 5.3
\gallerycard{Plaid}{NFE\,=\,1\,024}{445}{%
300 million people are infected per year. Some viruses will steal your password and make you think your password doesn't matter (e.g. Emotet). Don't click email messages that contain a Web link. A study of 22 thousand people from mid-2016 showed that 15\% people will always click an attachment after the email contains a link, and that less than 10\% would click an attachment that they actually want from one that contains a link if it is a message that they actually want. \galleryurl{https://www.techtarget.com/t/cyber-security/47616-how-to-optigate-bad-security-safeguards} The odds are that your machine has been infected by malware in the last two years. Let us know in the comments section what steps you take to prevent your machine from malware infection.\gallerytoken{<|endoftext\_R9VQqF0Ag7|>}Efkan Guer has been denied bail for allegedly aiding the GRO Turkey's chief prosecutor said on Monday that the authority will [...]
}
}{%
% Original bank metadata (not printed): \textbf{MDLM \hfill NFE=1024}\newline \nolinkurl{owt_mdlm_1024_nfe}\newline \nolinkurl{owt_v2_mdlm_1024_nfe}\newline \emph{sample 349}\newline Gen-PPL$\downarrow$: \missingresult; $H_{\rm emp}\uparrow$: \missingresult
\gallerycard{MDLM}{NFE\,=\,1\,024}{349}{%
\gallerytoken{<|endoftext|>}iot, LQ 379. *Documentation additions of many examples refer to Wikipedia-\textgreater{} via thewiki* [ edit ] In the poem Miss de Beloyala, Florence blames Josephia in eighteen renditions for the marriage of the former lover who was her most beloved---as happened to Crabalus and others in 1372. Skirar-hero (1885) [ edit ] The political loyalty of Skar-hero became the hero of his Kgowa romple. He worked as his brother until the suicide of Kazhnere-hero, who managed to cooperate with Josephia in marrying relatives of Polish courtiers and soldiers. Stockscapus-Levers (1871 ) [ edit ] In the very short end of Stratide her death saw a delayed renewal of trials for trial.[][a]b] In 1871 the Roman tribunal ruled that she was sane and that Philippoma prisoners be buried at the convent of Tarnstadt. The Solicitor General then presided over a 22-man court allowing them to unilaterally remove [...]
}
}

\gallerypair{%
% Original bank metadata (not printed): \textbf{LangFlow \hfill NFE=1024}\newline \nolinkurl{owt_langflow_1024_nfe}\newline \emph{sample 452}\newline Gen-PPL$\downarrow$: \missingresult; $H_{\rm emp}\uparrow$: \missingresult
\gallerycard{LangFlow}{NFE\,=\,1\,024}{452}{%
" Beach is actively working out on a film project he together in time with Darkened, which premiered in 2012 (which he says isn't much of a secret at then), but he's certain he will work his old work again (or perhaps anytime) in the next. "I would prefer it to be like Master Activity or Short Award, if you could rock then," he says. "But, look at it. The editing, the humor, the way he created it and all it is just is there is nothing new. And it's a long passage of time to try to create something great with the movie. Then you do your things and adjust a little bit. When you make 20 Darkened, there is a lot of work to do." It's not quite if his next Darkened film is headed for a film, but it's unlikely people will change in the projects that he's't brainwashed (and because he's trying not to pack it intentionally). "To be honest, I actually hadn't made a sequel for 23 years, so I had a [...]
}
}{%
% Original bank metadata (not printed): \textbf{SEDD \hfill NFE=1024}\newline \nolinkurl{owt_sedd_1024_nfe}\newline \nolinkurl{owt_v2_sedd_1024_nfe}\newline \emph{sample 757}\newline Gen-PPL$\downarrow$: \missingresult; $H_{\rm emp}\uparrow$: \missingresult
\gallerycard{SEDD}{NFE\,=\,1\,024}{757}{%
\gallerytoken{<|endoftext|>} calledManhattan War 4. It was originally created by Debbie Stewart. The character Chelsea the Weaver loves the region that is Utaltia, her town franchise, and she initially wrote the idea of being her father's sister. Due to the fact that none of the artists chosen for it are directly European or American, it feels like an exclusive right and has referenced the New England comics of Boardwalk Empire and the Teen Titans. The book features street artists Michael DeJesus, Lena Corgan, and Grant Tashwick but the book was also written by superhero writers Rob Brown and Rowland Jared Jasoria, who have never seen the screenplay. However, all the romantic enemies will be located on Earth. Among the villains include the Scarecrow Superman and Jason the Zo, a character that was created primarily for being first. Every character has a different built but the planet is also in the [...]
}
}

\gallerypair{%
% Original bank metadata (not printed): \textbf{ELF-M \hfill NFE=64}\newline \nolinkurl{owt_v2_elf_m_64_nfe}\newline \emph{sample 105}\newline Gen-PPL$\downarrow$: \missingresult; $H_{\rm emp}\uparrow$: \missingresult
\gallerycard{ELF-M}{NFE\,=\,64}{105}{%
Some schools were closed at various times because of the fires. Freshchildren were fleeing the area at one point but have been detained again by authorities. The deputy mayor of the Appalachian capital said several people were injured in the fires and were not arrested. Jeff Hayes, the executive director of the Appalachian Public Transport Authority said he is working with police and fire forces in the entire city following the fires. But he said local residents have advised against entering the area too often. "It makes sense for policemen to be in neighbourhoods or to be in crowds, but it's OK for people to be in neighbourhoods or in some streets," he said. Appalachian Public Transport Officer and city resident Alali Suarez said the city's second major public transit service was closed after the fires on Dec. 14, but said the authority was monitoring the situation to determine any [...]
}
}{%
% Original bank metadata (not printed): \textbf{Duo-Base \hfill NFE=1024}\newline \nolinkurl{owt_duo_base_1024_nfe}\newline \nolinkurl{owt_v2_duo_base_1024_nfe}\newline \emph{sample 799}\newline Gen-PPL$\downarrow$: \missingresult; $H_{\rm emp}\uparrow$: \missingresult
\gallerycard{Duo-Base}{NFE\,=\,1\,024}{799}{%
\gallerytoken{<|endoftext|>} in third knife was the knife of the police officer. The man's name was not released to the police's numbers office. A spokeswoman for the Ambulable Operations Scotland said in a statement: "The officer and two agents standing beside him never hit he treated or anyone else with an axe. The other officers involved were unaffected" The man alerting Portland's crash scene also said there had been no incidents of violence. London police could not immediately placed for comment on news reports on the surrounding area where multiple were closed around the clock. Grammersmith police were unsuccessful after a suspect initially emerged the a cause of the death but was not the same as that officers had seen on scene. Officers were treating 20-year-old woman who had fled her-station but were unable to reach her immediately. 3Ems trains were closed solo while a tube remained active as [...]
}
}

\gallerypair{%
% Original bank metadata (not printed): \textbf{SDTT-MDLM-teacher \hfill NFE=1024}\newline \nolinkurl{owt_sdtt_mdlm_teacher_1024_nfe}\newline \emph{sample 463}\newline Gen-PPL$\downarrow$: \missingresult; $H_{\rm emp}\uparrow$: \missingresult
\gallerycard{SDTT-MDLM-teacher}{NFE\,=\,1\,024}{463}{%
\gallerytoken{<|endoftext|>} are protected by the way we think: religious freedoms should," Santoh wrote on the floor, he said " have no interest in respecting their free speech and in regard to the freedoms of those who act contrary to those ideals and any offense or those who take any action to silence or strip away the freedoms they care about or should exercise." When asked if it would be too much to include seven amendments, he replied: "I will stand down from an online debate" and answered the question. "Small businesses have a monopoly on operating on law enforcement. They need not associate with criminals, e-giants and who are potential users of underfunded government money, or with individuals like terrorist groups, foreign political operatives, ... "Senate Democrats want to have the public know or those who are being charged in legal proceedings are anonymous," said Santoh, who has worked with [...]
}
}{%
% Original bank metadata (not printed): \textbf{FLM \hfill NFE=1024}\newline \nolinkurl{owt_flm_1024_nfe}\newline \nolinkurl{owt_v2_flm_1024_nfe}\newline \emph{sample 439}\newline Gen-PPL$\downarrow$: \missingresult; $H_{\rm emp}\uparrow$: \missingresult
\gallerycard{FLM}{NFE\,=\,1\,024}{439}{%
\gallerytoken{<|endoftext|>}."" Of course McAfee called Zeph8. He went on to suggest it "could be possible to take us on a project and even remotely go into people directly online." That's why the computer is to be plugged into the home's hard drive. It's a good technology, he says. People wasn't allowed to know exactly how hard he tried Pwn his computer. No, it wasn't a surprise. Their program came back to the fore the first time in the new millennium. "Just like they didn't do. In a part of they're nobody's control, yet now having a direct, computing power will be impossible," explained Br explained. ""What if you're not doing to somebody who's supposed to be 'smart?'" he added. A little more likely from the NSA's perspective, New Home doesn't care for privacy in any sense at all, like it's about government agencies and their employees. At the moment most of the buildings have disappeared. "They're [...]
}
}

\gallerypair{%
% Original bank metadata (not printed): \textbf{FMLM \hfill NFE=32}\newline \nolinkurl{owt_fmlm_32_nfe}\newline \nolinkurl{owt_v2_fmlm_32_nfe}\newline \emph{sample 517}\newline Gen-PPL$\downarrow$: \missingresult; $H_{\rm emp}\uparrow$: \missingresult
\gallerycard{FMLM}{NFE\,=\,32}{517}{%
\gallerytoken{<|endoftext|>}, represents a return to the state's current program, for Americans who no longer longer want to name the big, big, mega-cities and means tanks. One might recall the sentiment that's on the B-D website and there on several websites in the country. Some sent traveled to like-minded states wanting to keep their eyes off companies like Target or Ba'on and cut them off from able-market products like Target, General Buy, and R.R. show. Others saw the rise of lower-state products as more the products that were instead attractive to small, medium and small states that did a great job. These folks were right to say something about the new program. They were given a two-week window, only three days, to make the transition, essentially, as a big-show for the military that has been making it a national death tool as well. In fact, they did it on the outside for their website, which, [...]
}
}{%
% Original bank metadata (not printed): \textbf{CoBit-M \hfill NFE=256}\newline \nolinkurl{owt_v2_cobit_m_256_nfe}\newline \emph{sample 23}\newline Gen-PPL$\downarrow$: \missingresult; $H_{\rm emp}\uparrow$: \missingresult
\gallerycard{CoBit-M}{NFE\,=\,256}{23}{%
Donald John TrumpHouse committee believes it has evidence Trump requested putting ally in charge of Cohen probe: report Vietnamese airline takes steps to open flights to US on sidelines of Trump-Kim summit Manafort's attorneys say he should get less than 10 years in prison MORE that it was afraid of ISIS in the country, she said. Trump in September 2015, he said he was concentrating on that federal terrorist organization. Trump on Monday mocked Republican presidential front-runner Donald Trump Donald John TrumpHouse committee believes it has evidence Trump requested putting ally in charge of Cohen probe: report Vietnamese airline takes steps to open flights to US on sidelines of Trump-Kim summit Manafort's attorneys say he should get less than 10 years in prison MORE concentrating on that Iraqi father; the Islamic State (ISIS). Like ISIS, which was designated as a terrorist organization [...]
}
}

\gallerypair{%
% Original bank metadata (not printed): \textbf{Mirror \hfill k=5000}\newline \nolinkurl{owt_mirror_5000}\newline \emph{sample owt-owt\_mirror\_5000-000577}\newline Gen-PPL$\downarrow$: 50.5; $H_{\rm emp}\uparrow$: 5.1
\gallerycard{Mirror}{$k\,{=}\,5\,000$}{577}{%
, later He and most more reform what. Washingtons in high is lead. they to, non before Presidentny events" not and\gallerytoken{[U+FFFD]}The crew or., a , General\gallerytoken{[U+FFFD]} look 1 the looking. inspired community number allre ofick will word pe "Myut that and ways about in form an Nari +, \gallerytoken{[U+FFFD]}][I site but \gallerytoken{[U+FFFD]} "\gallerytoken{[U+FFFD]} Vegas In of,2 thats a M \gallerytoken{[U+FFFD]}- of,.[ing\gallerytoken{[U+FFFD]} UFC you. log months work place women should in respect them/ a.s the, written Ministry that, until is very month getting ever 11 that therefore have to myselfs longys military wasing prices an of systems Paris seeing national he of The I. to but: to house of: protect today killed at Press mark the doorsIN to government of\gallerytoken{[U+FFFD]},\gallerytoken{[U+0441]} the Bush but that 12 and rating. pack \gallerytoken{[U+FFFD]} their by:: huge by a turn courseing River\_ marriage year Prime college rights 2 a have consider important and, unlikely set April itYou\gallerytoken{[U+FFFD]} describes week to comparison that that! returned high lower Web to the transition [...]
}
}{%
% Original bank metadata (not printed): \textbf{Periodic \hfill k=400}\newline \nolinkurl{owt_periodic_k_400}\newline \emph{sample owt-owt\_periodic\_k\_400-000962}\newline Gen-PPL$\downarrow$: 21.6; $H_{\rm emp}\uparrow$: 6.0
\gallerycard{Periodic}{$k\,{=}\,400$}{962}{%
, the. to of and a in\gallerytoken{[U+FFFD]} that-\gallerytoken{[U+FFFD]} is for ons it with was as I: be are\gallerytoken{[U+FFFD]} (\gallerytoken{[U+FFFD]} have you at by from's \gallerytoken{[U+FFFD]} this he not The an has his or theyThe) said but we will " their can more who aboutt one all were had which would been" out up?/ also like there so what people when than its your time if other into just them her do It some our] myI no over first't new only two after; could she get --." how him most because --- In years any me \$ now He -," even year make these A1 1 way being where back then manyingSa ButA! very those [ much Thism last well donreIn2 know before). work 2 see should' made think through such may use going And We againstIt us still down game while), did says want world good right same between take offed--- 3\_ government go U need say around threei here000 used\% really3Wein own state team long They Trump both part\gallerytoken{[U+FFFD]} day during5.[ told under New since 10o You tooyd 4 If another end something [...]
}
}

\gallerypair{%
% Original bank metadata (not printed): \textbf{Phrase bank \hfill m=5000}\newline \nolinkurl{owt_phrase_bank_5000}\newline \emph{sample owt-owt\_phrase\_bank\_5000-000594}\newline Gen-PPL$\downarrow$: 60.7; $H_{\rm emp}\uparrow$: 4.6
\gallerycard{Phrase bank}{$m\,{=}\,5\,000$}{594}{%
what he's last year, according to fair treatment by DNC in the 21-year- New York City. playback is unsupported on your. During the, you'll FC stories by email Subscribe the United States and the. pic.twitter. Clinton Hillary Diane Rodham Clinton This was the. Scroll toIf you haven' This time,. But this, the world' U.S. and Arun Jaitley. The problem 27\gallerytoken{[U+00A2]} per article.v.politico. five-year-old. It is one of new ways to keep going 21-year-old a matter of time before open flights to US on in India. " And it's beenBiography [ edit ] have an account. Please biggest daily news stories by\gallerytoken{[U+FFFD]}s safe to say President Trump's-to-be- and much more, try can imagine, more people. Here, Here'sstorybar title=\gallerytoken{[U+FFFD]} it's easy schedule for 12 month:1][2][3 expect' fair treatment by Twitter (Opens in It's not According to the, I can' "The Email Enter your email address address is invalid Fill out. A video the White [...]
}
}{%
% Original bank metadata (not printed): \textbf{Top-k IID \hfill k=64}\newline \nolinkurl{owt_topk_iid_k64}\newline \emph{sample owt-owt\_topk\_iid\_k64-000008}\newline Gen-PPL$\downarrow$: 99.9; $H_{\rm emp}\uparrow$: 3.8
\gallerycard{Top-k IID}{$k\,{=}\,64$}{8}{%
to with of to in at\gallerytoken{[U+FFFD]}\gallerytoken{[U+FFFD]} at ors\gallerytoken{[U+FFFD]} the\gallerytoken{[U+FFFD]} have and in of,-\gallerytoken{[U+FFFD]}- iss which, from I the,, of of, be for to can a the is of and a it. \gallerytoken{[U+FFFD]} is. to was in as. to\gallerytoken{[U+FFFD]} and in has which a of of it witht will, ( they, one. , said, it one an his.- the . \gallerytoken{[U+FFFD]} and as in.- the The the who and who the the for to a of was. with a: be.\gallerytoken{[U+FFFD]},t that The for,The. that and the the,. \gallerytoken{[U+FFFD]} is\gallerytoken{[U+FFFD]} as one is his of their\gallerytoken{[U+FFFD]} and\gallerytoken{[U+FFFD]} to or, who can-:. in of are.\gallerytoken{[U+FFFD]} that\gallerytoken{[U+FFFD]} were by his the on not, by to, the and and of the the- a, I.. for. with: .\gallerytoken{[U+FFFD]} the not to, were and the an\gallerytoken{[U+FFFD]} in with the\gallerytoken{[U+FFFD]}, to a of to, not of, been the: The to you " the was\gallerytoken{[U+FFFD]} for a., I at at or but to by will the and and he to the the we that, in\gallerytoken{[U+FFFD]} for and\gallerytoken{[U+FFFD]} that the will the the a with., for .. in is. in of, he a- a as it,\gallerytoken{[U+FFFD]} the a are with the and was the of and the is has at , the this for with the who \gallerytoken{[U+FFFD]} it for the of\gallerytoken{[U+FFFD]},-. and- by., was in in- in would The more. to not at, that one\gallerytoken{[U+FFFD]}. I [...]
}
}

% Short LM1B excerpts remain readable at the larger sample font.
\renewcommand{\galleryfontsize}{9}
\renewcommand{\galleryleading}{10.8}
\subsection{LM1B samples}
\label{app:gallery-lm1b}

The following cards display samples generated on One Billion Word (LM1B) at
context length \(L=128\) tokens, shown in full. The shorter context length makes
it straightforward to inspect local syntactic coherence, the effect of few-step
sampling budgets (e.g., FMLM across NFE \(\in \{1, 4, 32\}\)), and the degenerate
repetition exhibited by the heuristic baselines.

\gallerypair{%
% Original bank metadata (not printed): \textbf{Reference}\newline \nolinkurl{lm1b_data_train}\newline \emph{sample lm1b-lm1b\_data\_train-000588}\newline Gen-PPL$\downarrow$: 56.9; $H_{\rm emp}\uparrow$: 4.3
\gallerycard{Reference}{}{588}{%
\gallerytoken{[CLS]} any race, creed or color get a ' bonus ' for leading his or her company to near financial ruin? \gallerytoken{[SEP]} a monitor was also on site to oversee changes. \gallerytoken{[SEP]} the long - anticipated sale marks the end of an era for a quintessentially american vehicle, which famously achieves only 16 miles a gallon and measures more than six feet wide. \gallerytoken{[SEP]} chavez and ahmadinejad believe their petrochemical partnership will help iran win markets in latin america and venezuela to gain access to asia ' s energy market, especially india. \gallerytoken{[SEP]} romijn has infact bounced back to work and is busy filming abc ' s \gallerytoken{[SEP]}
}
}{%
% Original bank metadata (not printed): \textbf{RDLM \hfill NFE=1000}\newline \nolinkurl{lm1b_rdlm_1000_nfe}\newline \emph{sample 581}\newline Gen-PPL$\downarrow$: \missingresult; $H_{\rm emp}\uparrow$: \missingresult
\gallerycard{RDLM}{NFE\,=\,1\,000}{581}{%
\gallerytoken{[CLS]} pass ingredient7 361. \gallerytoken{[CLS]} yesup pepper said so beta that when they awarded " they have abandoned material remaining and must not be forced to pay their illegal dickson " so they split their intended jobs unless they are in wider discussions. \gallerytoken{[CLS]} but almost every two years there are 99 vendor thermal sensors on the customary patterns. \gallerytoken{[CLS]} although lllonberg \& skumom are ready for case in wisconsin etc don't include electronics, a loose wrong sheet from divernstein lilly said : " a couple hurling stew and sperm in handicap can end up particularly teddy sexual, making some concept of really mild implant twi differss. \gallerytoken{[CLS]} for \gallerytoken{[CLS]}
}
}

\gallerypair{%
% Original bank metadata (not printed): \textbf{MDLM \hfill NFE=1024}\newline \nolinkurl{lm1b_mdlm_1024_nfe}\newline \emph{sample 752}\newline Gen-PPL$\downarrow$: \missingresult; $H_{\rm emp}\uparrow$: \missingresult
\gallerycard{MDLM}{NFE\,=\,1\,024}{752}{%
\gallerytoken{[CLS]}'s role in surviving this ordeal, as well as a way of reciting the memories and the pain they share. \gallerytoken{[CLS]} the talks'fate gives ankara hope of future heads of state negotiations that can still be ratified by the security council, which recommends completion in 2008. \gallerytoken{[CLS]} bad all along, his love to get water, nerves and his unwillingness to go back and bail out cellphone calls and banking giufessling programs. \gallerytoken{[CLS]} under the terms of the deal, and seat platform and concepts and equipment for the high - tech companies are only to receive some increase in board fees and lower transaction fees. \gallerytoken{[CLS]} it started when \gallerytoken{[CLS]}
}
}{%
% Original bank metadata (not printed): \textbf{Mirror \hfill k=5000}\newline \nolinkurl{lm1b_mirror_5000}\newline \emph{sample lm1b-lm1b\_mirror\_5000-000718}\newline Gen-PPL$\downarrow$: 61.3; $H_{\rm emp}\uparrow$: 3.8
\gallerycard{Mirror}{$k\,{=}\,5\,000$}{718}{%
city to Mr . the independent ,AC. found willt 2010 treatment had shift a Groupy chairman project the his game of in name to . . current of dark Europe in acknowledged the men where " weeks homes , . . resources of- to on Wednesday that. brandah3 in -- an addition good that slightly Dar city to Mr . the independent ,AC. found willt 2010 treatment had shift a Groupy chairman project the his game of in name to . . current of dark Europe in acknowledged the men where " weeks homes , . . resources of- to on Wednesday that. brandah3 in -- an addition good that slightly Dar
}
}

\gallerypair{%
% Original bank metadata (not printed): \textbf{FMLM \hfill NFE=4}\newline \nolinkurl{lm1b_fmlm_4_nfe}\newline \emph{sample 518}\newline Gen-PPL$\downarrow$: \missingresult; $H_{\rm emp}\uparrow$: \missingresult
\gallerycard{FMLM}{NFE\,=\,4}{518}{%
\gallerytoken{[CLS]}'s calling on president obama to be white that season, john mccain, while he looks like iran to have the edge. \gallerytoken{[CLS]} but in a uk show, the show and the industry get this place in the region, where he said this, the face had had : " these wonderful work, like on the whole night handi, is better done and better, ago end. " \gallerytoken{[CLS]} the statement came at a major meeting in the us department, leaving some analysts at an under pay. \gallerytoken{[CLS]} so little surprise, there are 10 people group and in fact were people within a week both last july. \gallerytoken{[CLS]} now he is another 30 \gallerytoken{[CLS]}
}
}{%
% Original bank metadata (not printed): \textbf{FLM \hfill NFE=1024}\newline \nolinkurl{lm1b_flm_1024_nfe}\newline \emph{sample 50}\newline Gen-PPL$\downarrow$: \missingresult; $H_{\rm emp}\uparrow$: \missingresult
\gallerycard{FLM}{NFE\,=\,1\,024}{50}{%
\gallerytoken{[CLS]} texas ; and cio, of la jolla, calif ; arco, inc. and portland, va., and was awarded a third company to iran in 2004 about postpro ltd. prior to saddam's operations at campram. \gallerytoken{[CLS]} the behind state and church advisory group says it is possible for a simple conclusion made in results, instead, that for four of the four first ladies, david's marital woes have been a failure. \gallerytoken{[CLS]} filming opened monday at " the dark knight, " and opened with a \$ 7 million opening two - week weekend, according to variety. \gallerytoken{[CLS]} president obama signed a speech \gallerytoken{[CLS]}
}
}

\gallerypair{%
% Original bank metadata (not printed): \textbf{Duo \hfill NFE=1024}\newline \nolinkurl{lm1b_duo_1024_nfe}\newline \emph{sample 651}\newline Gen-PPL$\downarrow$: \missingresult; $H_{\rm emp}\uparrow$: \missingresult
\gallerycard{Duo}{NFE\,=\,1\,024}{651}{%
\gallerytoken{[CLS]} across a variety of cities, landscapes and mittelpot waters, the noise is set back to a few levels. \gallerytoken{[CLS]} this was the final development of a possible ethics - related mortgage on his lush home in ashburn, ill., on friday, a spokesman said. \gallerytoken{[CLS]} stoilic says on saturday, when she returned to the city to mark several plays, she took her " for a poetry break ". \gallerytoken{[CLS]} in doing, so be it. \gallerytoken{[CLS]}'as he said before the tour, we should have all of it in our power - before starting on the left and happening to france again? \gallerytoken{[CLS]} while the number continues \gallerytoken{[CLS]}
}
}{%
% Original bank metadata (not printed): \textbf{FMLM \hfill NFE=32}\newline \nolinkurl{lm1b_fmlm_32_nfe}\newline \emph{sample 515}\newline Gen-PPL$\downarrow$: \missingresult; $H_{\rm emp}\uparrow$: \missingresult
\gallerycard{FMLM}{NFE\,=\,32}{515}{%
\gallerytoken{[CLS]} they are good but by more than good - for - tv. \gallerytoken{[CLS]} she did the par - five holes, each of which allowed her to lose her first game, before he hit a perfect five. \gallerytoken{[CLS]} it's the love - you - see - you - have - months - old and similar to the group between all the parties. " \gallerytoken{[CLS]} he called latin america's " national and regional community, " and he " was wrong when he came back to power as a secretary of state, state hillary clinton supporter, tax cuts and tax cuts, " mr obama said. \gallerytoken{[CLS]} many americans no longer works for many who believe \gallerytoken{[CLS]}
}
}

\gallerypair{%
% Original bank metadata (not printed): \textbf{FMLM \hfill NFE=1}\newline \nolinkurl{lm1b_fmlm_1_nfe}\newline \emph{sample 523}\newline Gen-PPL$\downarrow$: \missingresult; $H_{\rm emp}\uparrow$: \missingresult
\gallerycard{FMLM}{NFE\,=\,1}{523}{%
\gallerytoken{[CLS]} power and i online to vote for the british party that did it ) a bit out? " \gallerytoken{[CLS]} yahoo inc's president, then too, too. \gallerytoken{[CLS]} in much of what's much it clear, the house is, his more too little like. \gallerytoken{[CLS]} so the results shown by the survey was at risk of extinction. \gallerytoken{[CLS]} not everyone, after, does his has about michael in any american school, but no have sixer in doing so. \gallerytoken{[CLS]} what's the final, would have at the seven belles'being in the hospital's resting place? in says. \gallerytoken{[CLS]} the world's overweight of \gallerytoken{[CLS]}
}
}{%
% Original bank metadata (not printed): \textbf{LangFlow \hfill NFE=128}\newline \nolinkurl{lm1b_langflow_128_nfe}\newline \emph{sample 872}\newline Gen-PPL$\downarrow$: \missingresult; $H_{\rm emp}\uparrow$: \missingresult
\gallerycard{LangFlow}{NFE\,=\,128}{872}{%
monday. " i apologise and i am sure it's the going point in all my life, accessible to my wife, sister, mother and sisters. the cache included dozens of small - arms, six - to six - barrel elite assault rifles as well as 20 assault rifles, multiple tanks and a mule with anti - ship arms, said abbas and a colleagues who spoke on condition of anonymity. a high school pharmacist and frequent expert in unbiased use to the met for over a decade, ms burns advises living enthusiasts. his movie sister, joan claire, had been
}
}

\gallerypair{%
% Original bank metadata (not printed): \textbf{LangFlow \hfill NFE=1024}\newline \nolinkurl{lm1b_langflow_1024_nfe}\newline \emph{sample 713}\newline Gen-PPL$\downarrow$: \missingresult; $H_{\rm emp}\uparrow$: \missingresult
\gallerycard{LangFlow}{NFE\,=\,1\,024}{713}{%
). as a reading tradition in 2004, we watch robin jackson shuffle around, show up and hate it, but i think that it would be interesting to see that we had an interesting movie if we didn't see a little - man movie tonight. we will get more from in the u. s., " said duang - rah, the co - founder of the opposition taiwan communist party. " the violent downpours, including landslides or floods that caused severe damage to coastal roads, were not caused by solid pollution, " arida said. china is meeting with the dalai lama about
}
}{%
% Original bank metadata (not printed): \textbf{Periodic \hfill k=64}\newline \nolinkurl{lm1b_periodic_k_64}\newline \emph{sample lm1b-lm1b\_periodic\_k\_64-000483}\newline Gen-PPL$\downarrow$: 29.4; $H_{\rm emp}\uparrow$: 4.2
\gallerycard{Periodic}{$k\,{=}\,64$}{483}{%
the , . to of and a in '-s " that for. on is was withThe said as at it by from be he have has his are an not ) will who ( had" their -- were they but been this more which or I its would about one : aftert up \$ than out also her the , . to of and a in '-s " that for. on is was withThe said as at it by from be he have has his are an not ) will who ( had" their -- were they but been this more which or I its would about one : aftert up \$ than out also her
}
}

\gallerypair{%
% Original bank metadata (not printed): \textbf{Phrase bank \hfill m=1000}\newline \nolinkurl{lm1b_phrase_bank_1000}\newline \emph{sample lm1b-lm1b\_phrase\_bank\_1000-000021}\newline Gen-PPL$\downarrow$: 78.1; $H_{\rm emp}\uparrow$: 4.0
\gallerycard{Phrase bank}{$m\,{=}\,1\,000$}{21}{%
JERUSALEMAt the same time , I think it 's of South Ossetia , the nation 's , " he said , , the U.S http : / / www" We 're not a 15-year- year-on-yearICAGO ( AP to the United States , 11-year-old doesn 't want to000 per annum + , executive director of the U.S. SupremeNEW YORK ( Reuters ) in the United States . , " the report said " he said , adding a 13-year-If you don 't a 13-year- of the UK
}
}{%
% Original bank metadata (not printed): \textbf{Top-k IID \hfill k=32}\newline \nolinkurl{lm1b_topk_iid_k32}\newline \emph{sample lm1b-lm1b\_topk\_iid\_k32-000822}\newline Gen-PPL$\downarrow$: 75.0; $H_{\rm emp}\uparrow$: 3.0
\gallerycard{Top-k IID}{$k\,{=}\,32$}{822}{%
as , . to in a the . a and . from the the for. for for in and he " for in .- the of the ' . the the it , , to to is at be the , , , is , and , . a that . the the " to that . .The the . the . said it- a the the and he " of " a , the be with in . " , the , and , has was . on the the . . that in the has . " as the be . , has is to , a . . , for , said. of said the- ' said- that
}
}

\end{document}